\documentclass[letterpaper]{article} 
\usepackage{aaai2026}  
\usepackage{times}  
\usepackage{helvet}  
\usepackage{courier}  
\usepackage[hyphens]{url}  
\usepackage{graphicx} 
\usepackage{natbib}  
\usepackage{caption} 
\usepackage{algorithm}
\usepackage{algorithmic}

\usepackage{newfloat}
\usepackage{listings}
\DeclareCaptionStyle{ruled}{labelfont=normalfont,labelsep=colon,strut=off} 
\floatstyle{ruled}
\newfloat{listing}{tb}{lst}{}
\floatname{listing}{Listing}
\usepackage{booktabs}       
\usepackage{nicefrac}       
\usepackage{microtype}      
\usepackage{amsmath,amsfonts}
\usepackage{amssymb}
\usepackage{multirow}
\usepackage{subcaption}
\usepackage{xcolor}         
\usepackage{verbatim}

\newcommand{\etal}{\emph{et al.}} 
\newcommand{\Best}[1]{\textcolor{red}{\textbf{#1}}}
\newcommand{\SecondBest}[1]{\textcolor{cyan}{#1}}

\newcommand{\tao}[1]{\textcolor{black}{#1}}
\newcommand{\taore}[1]{\textcolor{black}{#1}}
\newcommand{\lyw}[1]{\textcolor{black}{#1}}
\newcommand{\lywcr}[1]{\textcolor{black}{#1}}

\title{MVGD-Net: A Novel Motion-aware Video Glass Surface Detection Method}

\author {
    Yiwei Lu\equalcontrib,
    Hao Huang\equalcontrib,
    Tao Yan\thanks{Tao Yan is the corresponding author.}
}
\affiliations {
    School of Artificial Intelligence and Computer Science, Jiangnan University, China\\
    \{6243112037, 6223110021\}@stu.jiangnan.edu.cn, yantao.ustc@gmail.com
}

\begin{document}

\maketitle

\begin{abstract}
Glass surface ubiquitous in both daily life and professional environments presents a potential threat to vision-based systems, such as robot and drone navigation. To solve this challenge, most recent studies have shown significant interest in Video Glass Surface Detection (VGSD). We observe that objects in the reflection (or transmission) layer appear farther from the glass surfaces. Consequently, in video motion scenarios, the notable reflected (or transmitted) objects on the glass surface move slower than objects in non-glass regions within the same spatial plane, and this motion inconsistency can effectively reveal the presence of glass surfaces. Based on this observation, we propose a novel network, named MVGD-Net, for detecting glass surfaces in videos by leveraging motion inconsistency cues. Our MVGD-Net features three novel modules: the Cross-scale Multimodal Fusion Module (CMFM) that integrates extracted spatial features and estimated optical flow maps, the History Guided Attention Module (HGAM) and Temporal Cross Attention Module (TCAM), both of which further enhances temporal features. A Temporal-Spatial Decoder (TSD) is also introduced to fuse the spatial and temporal features for generating the glass region mask. Furthermore, for learning our network, we also propose a large-scale dataset, which comprises 312 diverse glass scenarios with a total of 19,268 frames. Extensive experiments demonstrate that our MVGD-Net outperforms relevant state-of-the-art methods.
\end{abstract}

\begin{links}
    \link{Code and Datasets}{https://github.com/YT3DVision/MVGDNet}
\end{links}

\section{Instruction}

Glass surfaces, such as \tao{glass} windows, walls and doors, are \tao{ubiquitous} in our everyday lives. They are \tao{always transparent} and colorless, posing significant challenges for computer vision tasks, such as \tao{robot and drone navigation}~\cite{li2024hydra}, depth estimation~\cite{He2023depth} and 3D reconstruction~\cite{li2020TranReNet}. 
Previous methods have explored \tao{various} prior \tao{cues} or assumptions for single-image glass surface detection (GSD), such as contrasted context features~\cite{mei2020GDNet}, boundary cues~\cite{he2021EBLNet}, reflection phenomena~\cite{lin2021GSDNet}, ghosting effect~\cite{yan2025GhostingNet}, semantic relation~\cite{lin2022RGBS}, and visual blurriness~\cite{qi2024VBNet}. 
\tao{Multimodal image-based GSD} methods have also explored RGB-Depth~\cite{lin2025DGSDNet}, Polarization~\cite{mei2022RGBP}, RGB-Thermal~\cite{huo2023GETR} and RGB-NIR~\cite{yan2024NRGlassNet} \tao{for more effective \tao{and robust} GSD}. 

However, real-world applications like robotic navigation and autonomous driving are video-centric rather than image-centric, and \tao{above} methods are tailored for \tao{single} images, \tao{failing to leverage temporal information}. Thus, \tao{most} recently, Liu \etal~\cite{liu2024VGSDNet} proposed the first VGSD method, named VGSD-Net, which leverages temporal information in videos to improve detection. \tao{Though} VGSD-Net exploits reflections \tao{from each frame} to \tao{aid} GSD, it may fail to obtain accurate reflections in challenging \tao{scenes} (as shown in Fig.~\ref{fig:instrction_visual_comparison}) due to the absence of ground-truth reflection supervision.

\newlength{\newsubwidth}
\setlength{\newsubwidth}{0.24\linewidth}
\begin{figure}[t]

	\renewcommand{\tabcolsep}{0.8pt}
	\renewcommand\arraystretch{0.6}
        \centering
            \begin{tabular}{cccc}
            
                \includegraphics[width=\newsubwidth]{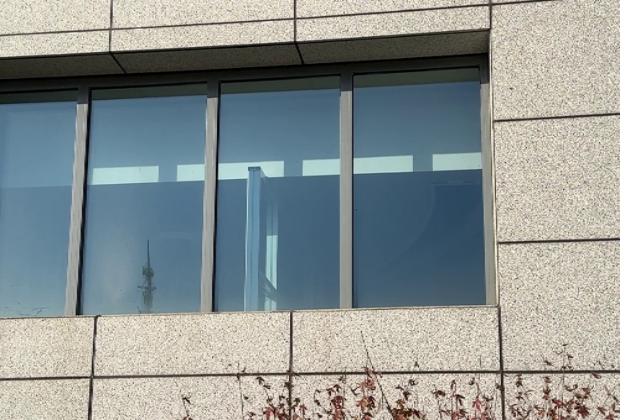}&
                \includegraphics[width=\newsubwidth]{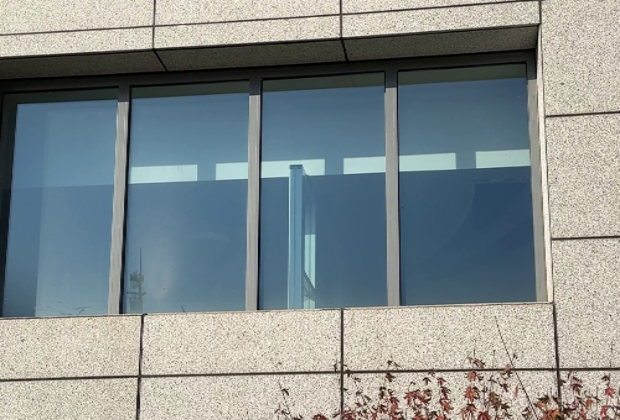}&
                \includegraphics[width=\newsubwidth]{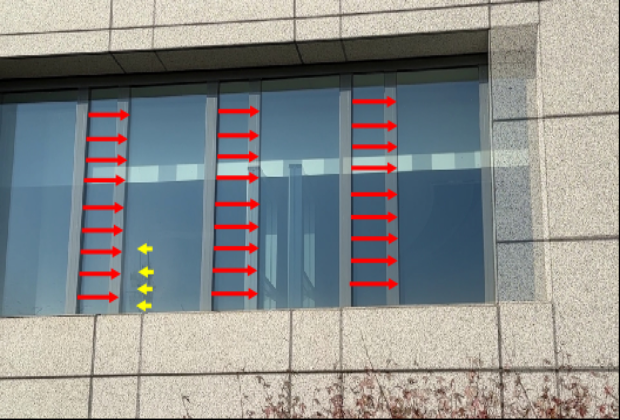}&
                \includegraphics[width=\newsubwidth]{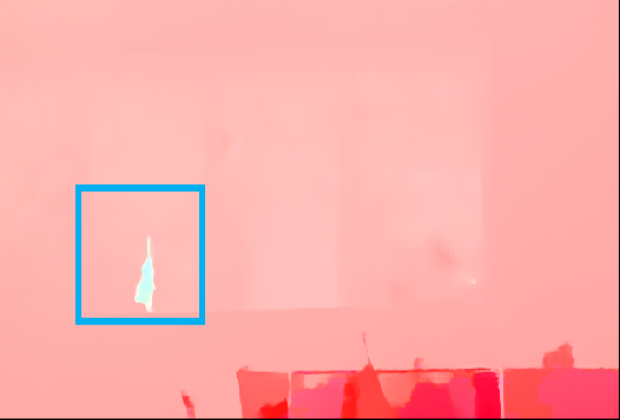}\\ 
  
                \includegraphics[width=\newsubwidth]{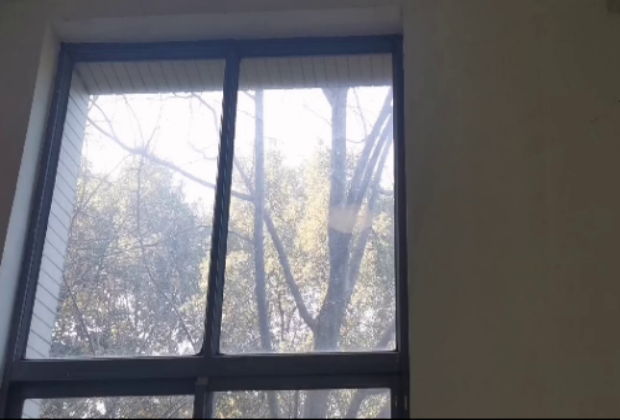}&
                \includegraphics[width=\newsubwidth]{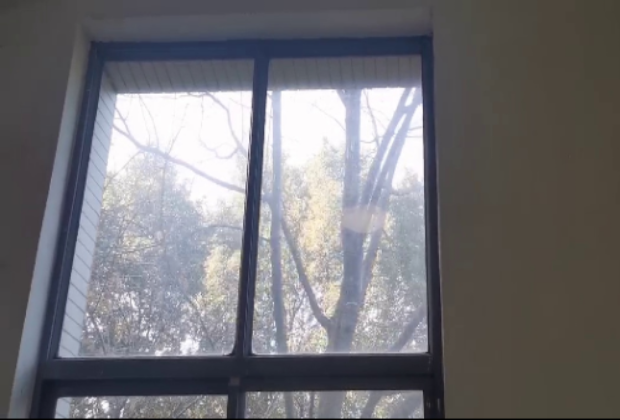}&
                \includegraphics[width=\newsubwidth]{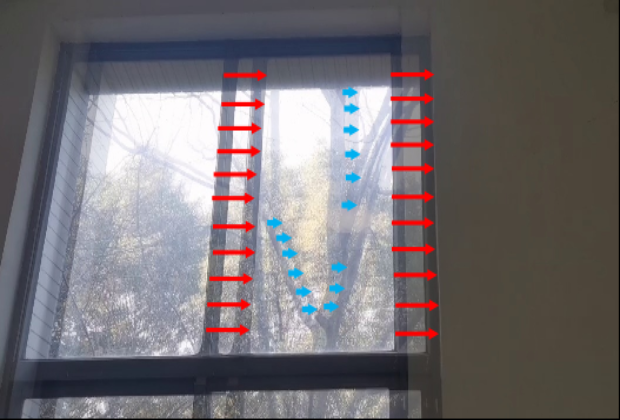}&
                \includegraphics[width=\newsubwidth]{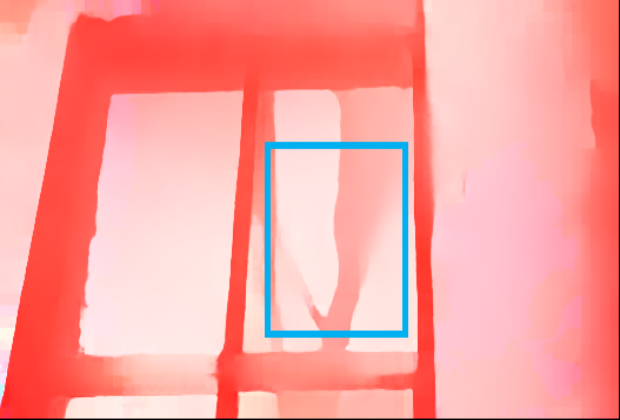}\\  

               \fontsize{9.0pt}{\baselineskip}\selectfont{Frame $t$}&
                \fontsize{9.0pt}{\baselineskip}\selectfont{Frame $t+10$}&
                \fontsize{9.0pt}{\baselineskip}\selectfont{Motion Cue}&
                \fontsize{9.0pt}{\baselineskip}\selectfont{Optical Flow}\\
                
            \end{tabular}
        \caption{Examples of the motion inconsistency in indoor and outdoor scenes. The $1st$ and $2nd$ columns show the frames at time $t$ and time $t+10$, respectively. The $3rd$ column shows the inconsistent motion cues \tao{on glass surfaces}, indicated by arrows. The $4th$ column shows the optical flow maps.}
        \label{fig:instrction_motion_cue}
\end{figure}

Neuroscience studies \tao{demonstrate} that humans rely on dynamic perceptual cues to identify glass regions in daily life. \tao{The references}~\cite{tamura2016PSMG,tamura2017MC4MG,tamura2018DC4MG} reveal that rotational, parallax, forward, and backward motions can induce motion inconsistencies, which are useful cues for GSD.
In \tao{real-world scenarios, we have observed that objects in the reflection (or transmission) layer appear farther from the glass surfaces. Consequently, in video motion scenarios, the notable reflected (or transmitted) objects on the glass surface move more slowly than objects in non-glass regions within the same spatial plane.}
\lyw{As shown in the \tao{$1st$ scene} of  Fig.~\ref{fig:instrction_motion_cue},} reflection moves more slowly than non-glass regions when the camera moves in front of glass surfaces. These inconsistent \tao{motion} cues imply the \tao{existence} of reflections, which could be further exploited for GSD. \lyw{\tao{Moreover}, as illustrated in the \tao{$2nd$ scene} of Fig.~\ref{fig:instrction_motion_cue}, even in indoor scenes with weak reflections, motion inconsistency can still arise due to transmitted objects being located at greater depths.}
\lyw{Unlike Warren \etal~\cite{warren2024MGVMD}’s \tao{mirror detection method exploiting motion inconsistency}, our method further accounts for special cases \tao{of GSD}, such as open doors and windows, by applying the \tao{primary} mask to 
\taore{filter inconsistent motion cues from obvious non-glass regions.}}
\tao{Specifically, we adopt RAFT~\cite{teed2020raft} to calculate optical flow maps, which can reveal the potential \tao{location} of glass surfaces, as shown in Fig.~\ref{fig:instrction_visual_comparison}.}

\setlength{\newsubwidth}{0.158\linewidth}
\begin{figure}[t]
	\renewcommand{\tabcolsep}{0.8pt}
	\renewcommand\arraystretch{0.6}
        \centering
            \begin{tabular}{cccccc}
                \includegraphics[width=\newsubwidth]{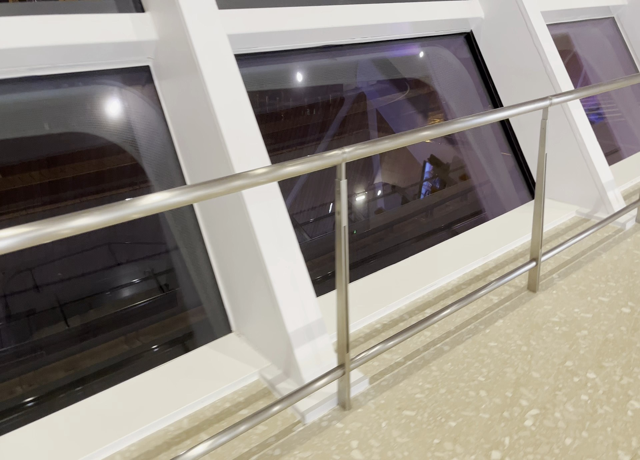}&
                \includegraphics[width=\newsubwidth]{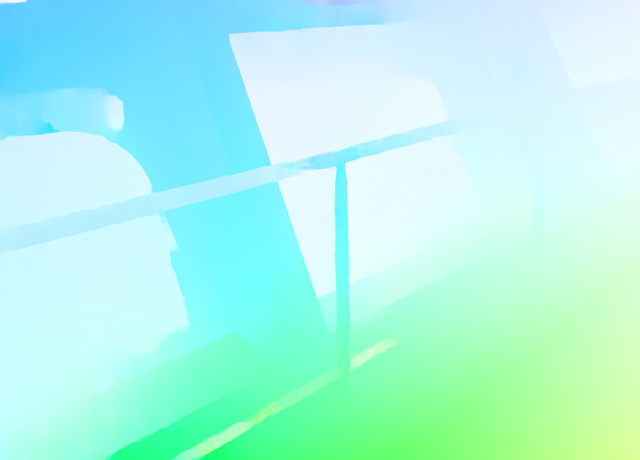}&
                \includegraphics[width=\newsubwidth]{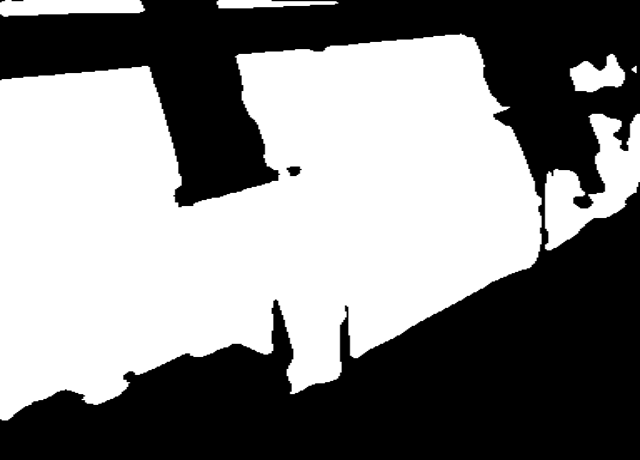}&
                \includegraphics[width=\newsubwidth]{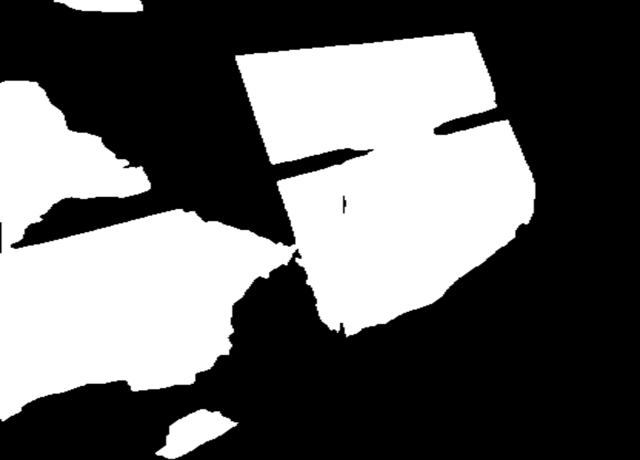}&
                \includegraphics[width=\newsubwidth]{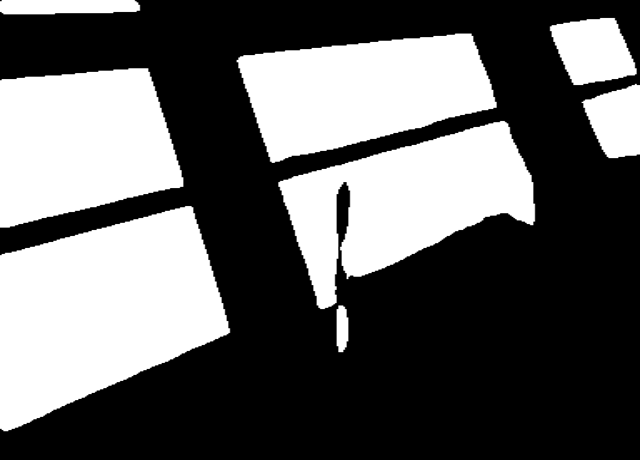}&
                \includegraphics[width=\newsubwidth]{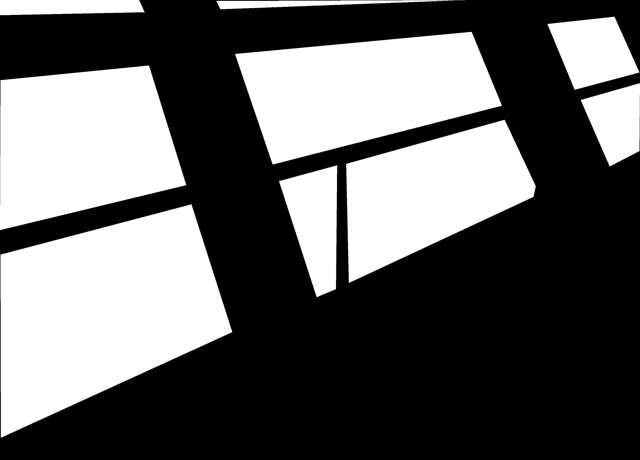}\\ 
  
                \includegraphics[width=\newsubwidth]{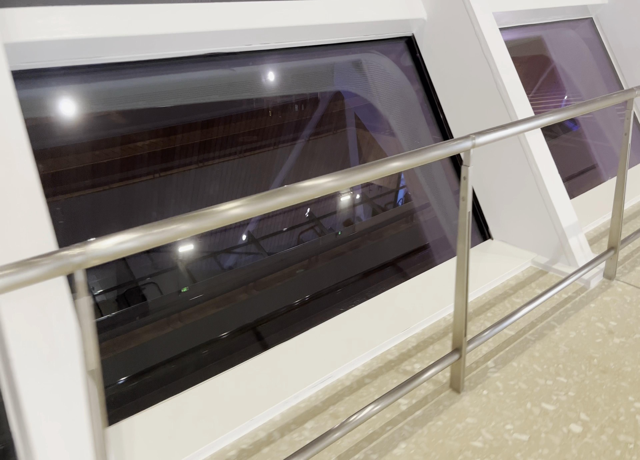}&
                \includegraphics[width=\newsubwidth]{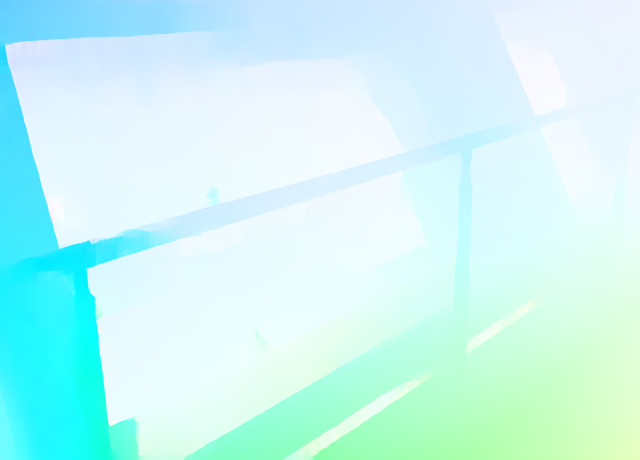}&
                \includegraphics[width=\newsubwidth]{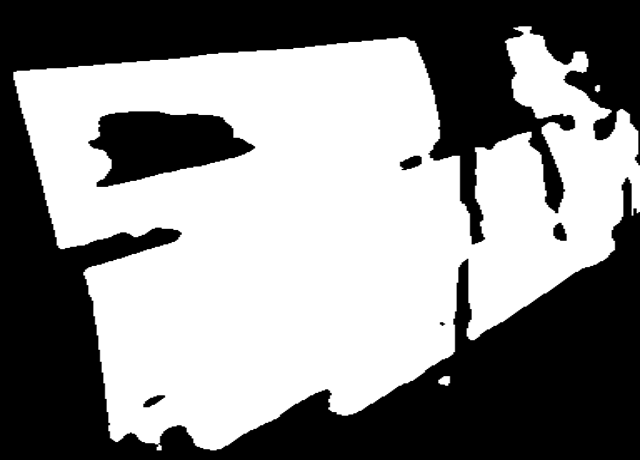}&
                \includegraphics[width=\newsubwidth]{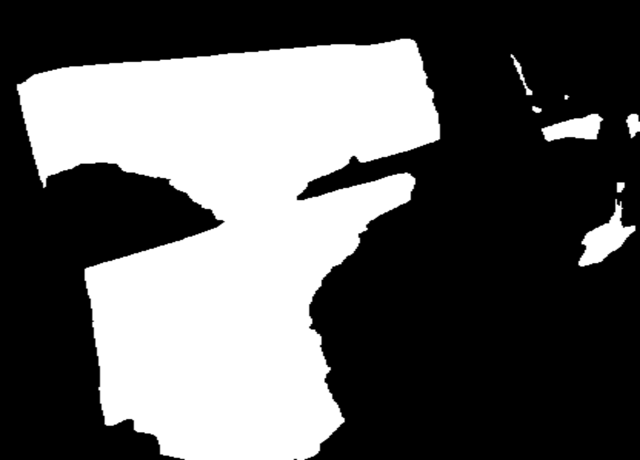}&
                \includegraphics[width=\newsubwidth]{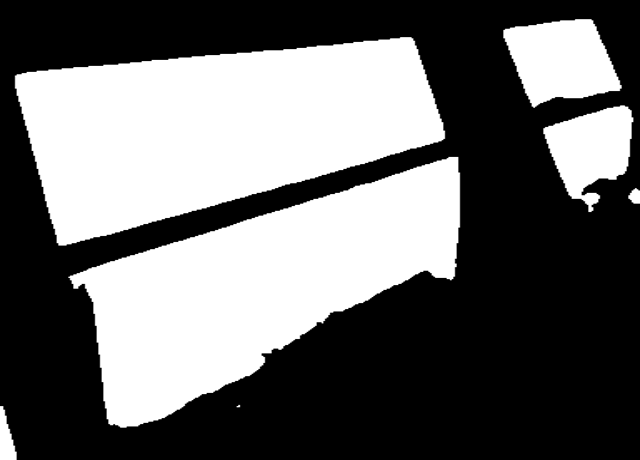}&
                \includegraphics[width=\newsubwidth]{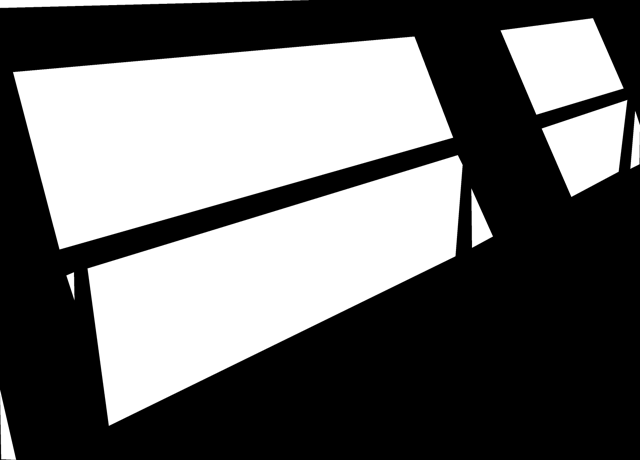}\\

               \fontsize{9.0pt}{\baselineskip}\selectfont{Frames}&
                \fontsize{9.0pt}{\baselineskip}\selectfont{Flow Map}&
                \fontsize{9.0pt}{\baselineskip}\selectfont{GSDNet}&
                \fontsize{9.0pt}{\baselineskip}\selectfont{VGSD-Net}&
                \fontsize{9.0pt}{\baselineskip}\selectfont{Ours}&
                \fontsize{9.0pt}{\baselineskip}\selectfont{GT}\\
                
            \end{tabular}
        \caption{Existing methods may \tao{under/over-}detect glass surfaces in challenge scenes.
        Our method utilizes motion inconsistency \tao{cues} to guide GSD and outperforms \taore{competitors}.}
        \label{fig:instrction_visual_comparison}
\end{figure}

In this paper, we propose the \textbf{M}otion-aware \textbf{V}ideo \textbf{G}lass Surface \textbf{D}etection \textbf{Net}work (MVGD-Net), comprising three key modules: (1) the Cross-scale Multimodal Fusion Module (CMFM), which integrates features from two modalities across multiple scales; (2) the History-Guided Attention Module (HGAM), which leverages historical information to refine current frame predictions; and (3) the Temporal-Spatial Decoder (TSD), which balances the fusion of temporal and spatial features in the decoding process. To train our network, we also construct a large-scale dataset, MVGD-D, comprising $19,268$ frames across $312$ videos, with manually annotated glass surface masks for each frame. Our main contributions are summarized as follows:
\begin{itemize}

    \item We propose a novel network, \tao{named} MVGD-Net, which exploits motion inconsistency \tao{from} video \tao{sequence} for VGSD task. \tao{MVGD-Net features three novel modules:} the CMFM for cross-scale multimodal fusion, the HGAM for \tao{exploiting} temporal context, and the TSD for effectively balancing temporal and spatial feature integration.

    \item We \tao{propose the} MVGD-D \tao{dataset}, a large-scale dataset containing $312$ videos with a total of $19,268$ frames, each with corresponding manual annotations, \tao{for GSD}.

    \item Extensive experiments demonstrate \tao{the outperformance of our proposed} method \tao{compared with the} relevant state-of-the-art methods.

\end{itemize}

\section{Related Work}
\paragraph{Glass Surface Detection (GSD).} Early works primarily focused on leveraging single-frame visual cues, such as contrasted contextual features~\cite{mei2020GDNet}, reflection priors~\cite{lin2021GSDNet}, boundary detection~\cite{he2021EBLNet,Fan2023RFENet}, semantic correlations~\cite{lin2022RGBS}, blurry effects~\cite{qi2024VBNet}, and ghosting cues~\cite{yan2025GhostingNet}. 
To capture more comprehensive scene information, recent methods have explored multi-modal imaging for GSD. For example, RGB-polarization~\cite{mei2022RGBP}, RGB-thermal~\cite{huo2023GETR}, RGB-NIR~\cite{yan2024NRGlassNet}, and RGB-D~\cite{lin2025DGSDNet} approaches leverage additional inputs to provide richer and more distinctive glass cues compared to RGB-only methods.

Despite these advancements, all the above methods are designed for \tao{singles} images and fail to utilize the temporal information present in video sequences.

\paragraph{Video Salient Object Detection (VSOD).} Video Salient Object Detection aims to identify the most visually significant objects across video frames by integrating spatial and temporal information. To improve detection accuracy, early works have utilized motion cues~\cite{wang2017video} and optical flow maps~\cite{li2019motion} to localize salient objects. More recent approaches~\cite{oh2019video, cheng2021rethinking, cheng2024putting} further enhance temporal consistency by introducing memory modules.
However, these methods often detect the objects behind the glass rather than the glass surface itself, due to the glass region may not always represent the most visually significant objects within a scene.

\begin{figure*}[ht]
\centering
\includegraphics[width=0.998\textwidth]{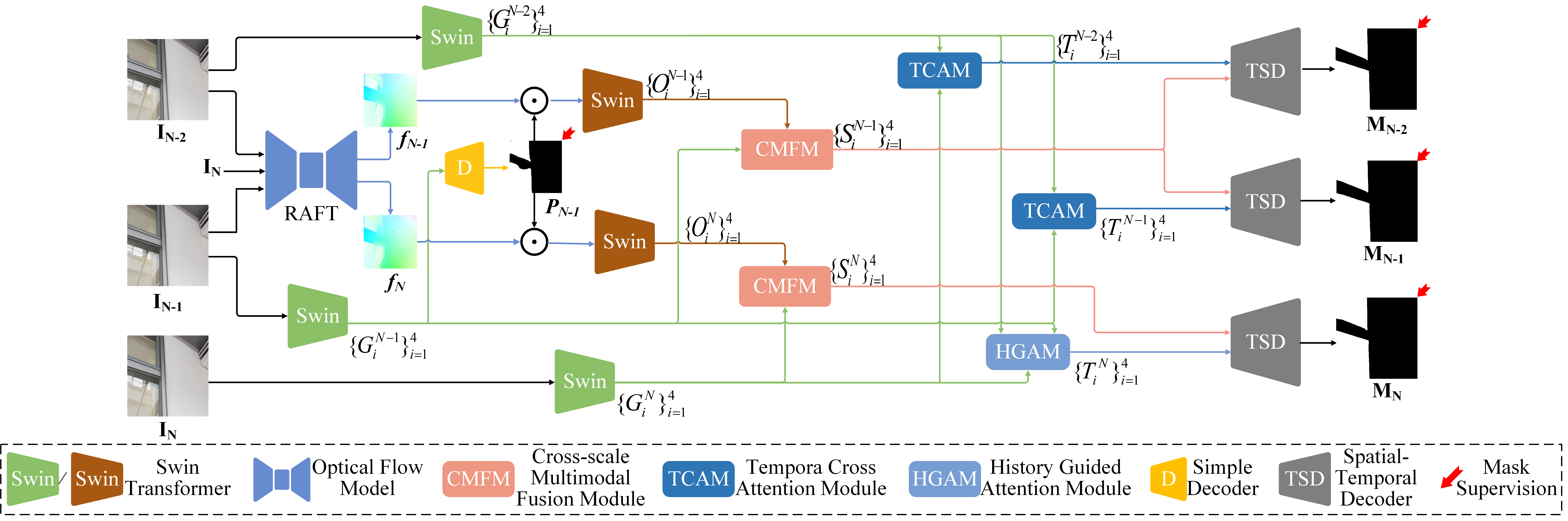}
\caption{The \tao{structure} of our proposed \textbf{M}otion-aware \textbf{V}ideo \textbf{G}lass Surface \textbf{D}etection \textbf{Net}work \tao{(MVGD-Net)}.}
\label{fig:MVGD-NET}
\end{figure*}

\section{Our Method}

\subsection{Pipeline}

Our key observation is that glass \tao{surfaces always show} motion inconsistency \tao{compared to non-glass regions located in the same position. This phenomenon arises from the varied depth perception created by the reflection (or transmission) layer.}
This observation motivates us to fully exploit \tao{motion inconsistency} contained within optical flow maps to \tao{aid VGSD}.
\tao{The structure of our proposed MVGD-Net is} shown in Fig.~\ref{fig:MVGD-NET}. 

Specifically, \tao{three adjacent frames ($I_{N-2}$, $I_{N-1}$, $I_{N}$)} are first input into RAFT~\cite{teed2020raft} to estimate optical flow maps ($f_{N-1}$ and $f_{N}$), while simultaneously passing through \lywcr{a weight-shared} 
Swin backbone for feature \tao{extraction}. A coarse glass surface mask $P_{N-1}$ is generated by direct decoding of the intermediate features ${\{G_{i}^{N-1}}\}_{i=1}^4$, which \tao{is then used to exclude inconsistent motion cues from non-glass areas}. Following feature extraction via \lywcr{another weight-shared} Swin transformer, the optical flow features (${\{O_{i}^{N-1}}\}_{i=1}^4$, ${\{O_{i}^{N}}\}_{i=1}^4$) and RGB features (${\{G_{i}^{N-2}}\}_{i=1}^4$, ${\{G_{i}^{N-1}}\}_{i=1}^4$, ${\{G_{i}^{N}}\}_{i=1}^4$) are \tao{fed} into the CMFM module to extract spatial features (${\{S_{i}^{N-1}}\}_{i=1}^4$, ${\{S_{i}^{N}}\}_{i=1}^4$). \tao{At the same time}, RGB features are passed through the TCAM and HGAM modules to capture temporal features (${\{T_{i}^{N-2}}\}_{i=1}^4$, ${\{T_{i}^{N-1}}\}_{i=1}^4$, ${\{T_{i}^{N}}\}_{i=1}^4$). 
\taore{HGAM and TCAM aggregate temporal information across frames, stabilizing the model against local unreliable flow cues.}
Finally, the TSD module fuses the spatial features \tao{and} the temporal features to \tao{produce} the final prediction of the glass surface masks ($M_{N-2}$, $M_{N-1}$, $M_{N}$).

\subsection{Cross-scale Multimodal Fusion Module (CMFM)}
Camera shake and other factors can cause significant variations in motion between adjacent frames, so we estimate optical flow between adjacent frames to improve accuracy, where $f_{N-1} = \text{RAFT}(I_{N-2}, I_{N-1})$ and $f_{N} = \text{RAFT}(I_{N-1}, I_N)$.
The optical flow maps $f_{N-1}$ and $f_{N}$, refined by the glass mask $P_{N-1}$, can indicate the potential \tao{locations} of the glass surface\tao{s}. After being encoded by a Swin-Transformer~\cite{liu2022Swin}, the four-scale optical flow features ${\{O_{i}^{t}}\}_{i=1}^4$ and RGB features ${\{G_{i}^{t}}\}_{i=1}^4$ are fed into the CMFM to generate spatial features ${\{S_{i}^{t}}\}_{i=1}^4$, where $t \in \{N-1, N\}$. 

\begin{figure}[ht]
\centerline{\includegraphics[width=0.5\textwidth]{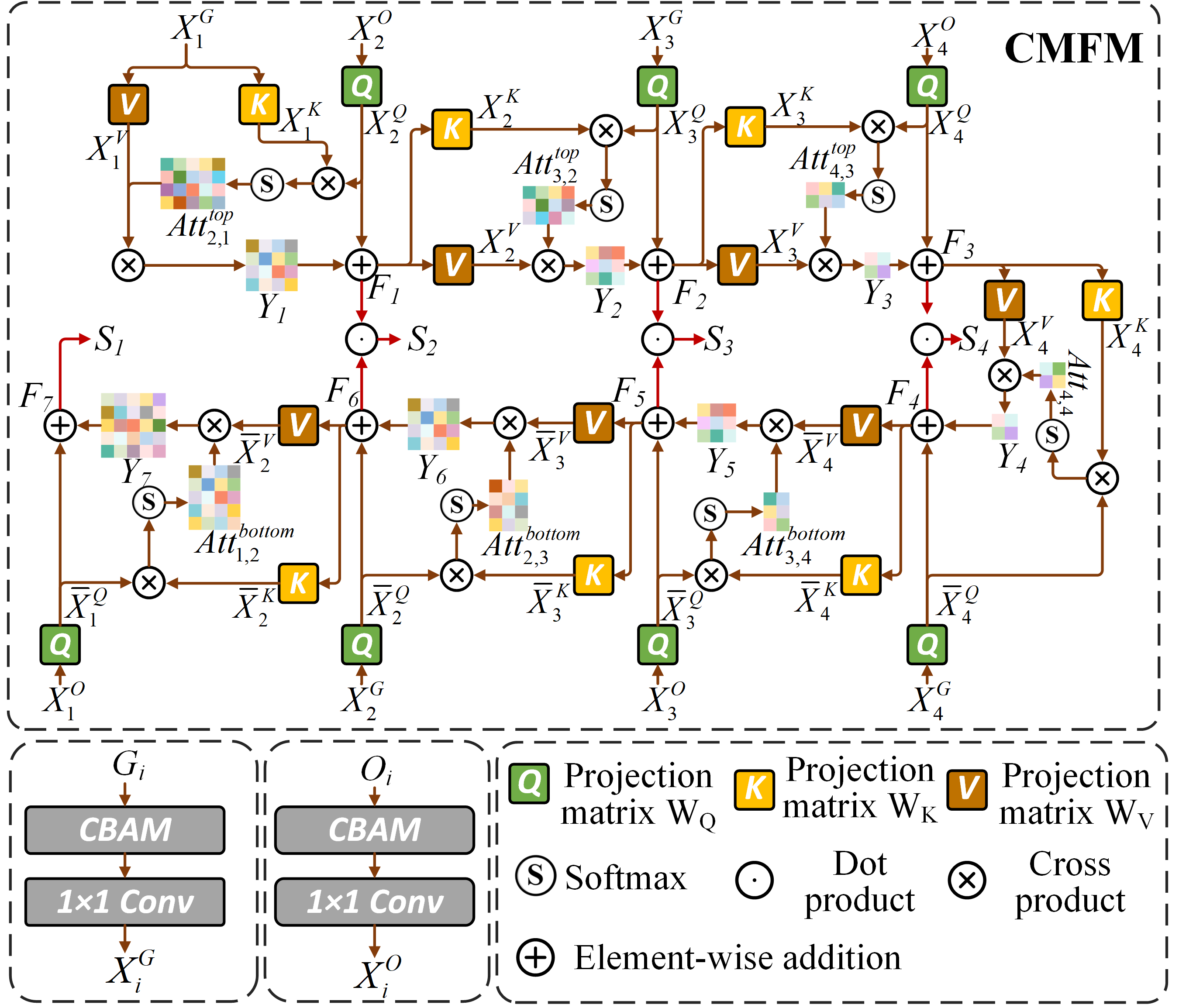}}
\caption{The structure of our CMFM.}
\label{fig:CMFM}
\end{figure}

The CMFM shown in Fig.~\ref{fig:CMFM} leverages the cues provided by optical flow to guide the learning of the GSD. 
The input features ${\{G_{i}}\}_{i=1}^4$ and ${\{O_{i}}\}_{i=1}^4$ are \tao{first} all projected to the channel \tao{size} $C_1$ to reduce the computational cost, \tao{as follows}:
\begin{equation}
    X_i^G=\text{Conv}(\text{CBAM}(G_i)), 
\end{equation}
where CBAM(·)~\cite{CBAM} is a lightweight attention module for feature refinement, and Conv(·) denotes a $1\times 1$ convolution used for channel dimension reduction. \tao{$C_1$ is set to $128$}.
Similarly, $X_i^O$ is \tao{generated from $O_i$} in the same way.


CMFM consists of seven cross-scale cross-attention blocks, each with independent parameter matrices. The fusion of all features is achieved through a U-shaped loop that proceeds from left to right \tao{(top branch)}, and then from right to left \tao{(bottom branch)}.
At each attention block, the new query features iteratively update the memory representations, enabling the progressive fusion of all eight feature maps.

In the left-to-right process of CMFM (\tao{top branch} of Fig.~\ref{fig:CMFM}), the feature maps are gradually downsampled, which can be regarded as a process of compressing spatial features to extract more effective representations. 
In the $i$-th attention block, the corresponding query, key, and value feature vectors are obtained via the projection matrices $W^Q_i$, $W^K_i$, and $W^V_i$, respectively. The feature vector $X_i^K$ is then transposed using a reshape operation.
\tao{While $i=1$, $X_i^K=W^K_1(X_1^G)$, $X_i^V=W^V_1(X_1^G)$, and $X_{i+1}^Q=W^Q_1(X_2^O)$.}
The attention map $Att_{i+1,i}^{top}$ is then \tao{calculated} as follow:
\begin{equation}
    Att_{i+1,i}^{top}=\mathrm{SoftMax}(X_{i+1}^Q\otimes X_i^K), 
\end{equation}
where $i \in [1, 3]$, and $\mathrm{SoftMax}(\cdot)$ denotes the normalization operation that maps values to the range $[0, 1]$. The attention map $Att_{i+1,i }^{top}$ is then applied to the memory ($X_i^V$) to obtain the fused features $Y_i \in \mathbb{R}^{H_{i+1}W_{i+1} \times C_1}$, as follows: 
\begin{equation}
    Y_i = Att_{i+1,i}^{top} \otimes X_i^V,    
\end{equation}
Subsequently, the query $X_{i+1}^Q$ is concatenated with the fused features:
\begin{equation}
    F_i = \text{R}(\text{MLP}(\text{LN}(Y_i)))+X_{i+1}^Q,
\end{equation}
where $\mathrm{MLP}(\cdot)$ \tao{represents the function of} multi-layer perceptron, \tao{and} $\mathrm{LN}(\cdot)$ represents layer normalization. $\mathrm{R}(\cdot)$ \tao{is the function of} reshape operation. 
%
\tao{Particularly}, \tao{while} $i = 4$, $X^O_4$ and $X^G_4$ are serving as the connection point between the top and bottom branches.

In the right-to-left process of CMFM (bottom \tao{branch of CMFM}), the feature maps are gradually upsampled, which can be regarded as a progressive enhancement of the \tao{important} features. The attention map $Att_{j-1,j}^{bottom}$, as well as the computation of $Y_i$ and $F_i$, follows a similar procedure to that described above. Here, $j \in {\{4, 3, 2}\}$ and $i = 9 - j$.

Finally, the features $F_i$ at the same scale are paired and fused via element-wise multiplication to obtain the final output features ${\{S_{i}}\}_{i=1}^4$. This process can be formulated as:
\begin{equation}
    S_i = 
    \begin{cases}
        F_7  &i=1, \\
        F_{i-1}\odot F_{8-i} &i=2,3,4.
    \end{cases}
\end{equation}



\begin{figure}[t]
\centerline{\includegraphics[width=0.41\textwidth]{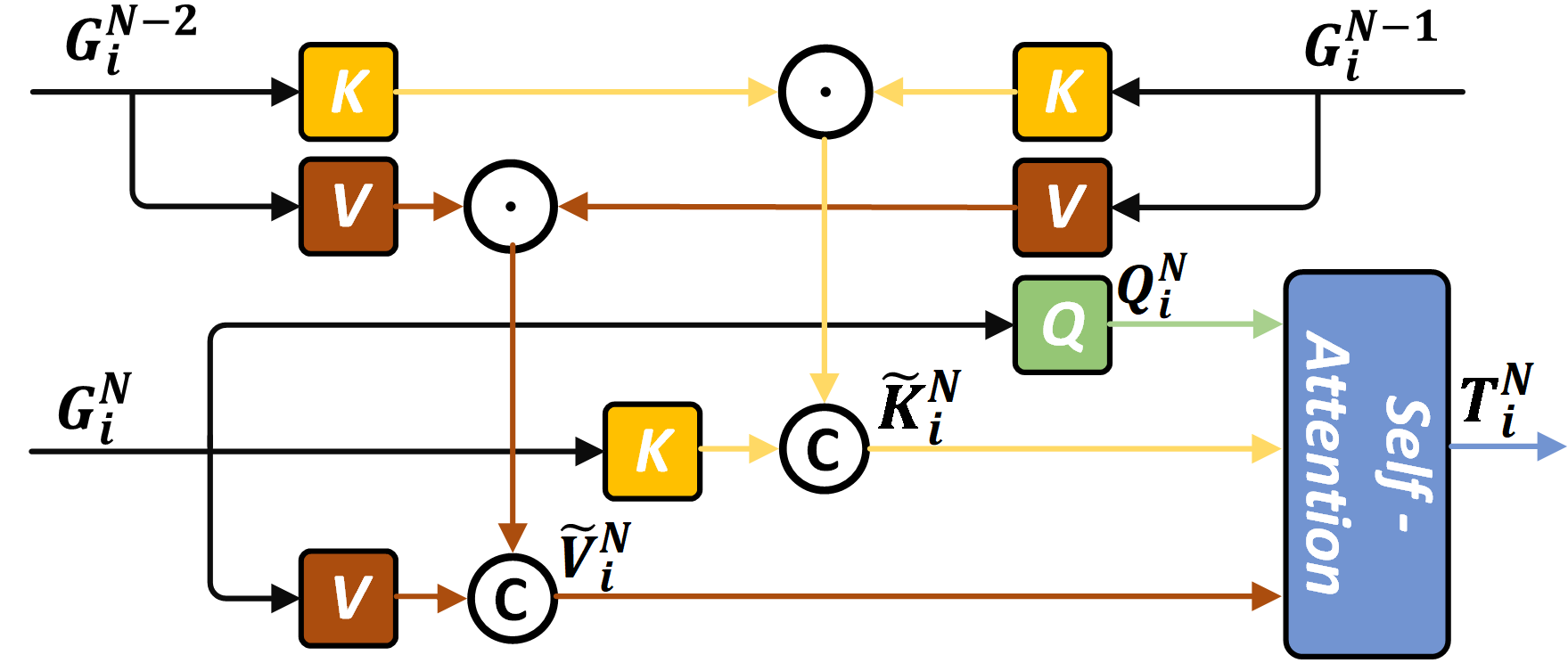}}
\caption{The structure of our HGAM.}
\label{fig:HGAM}
\end{figure}

\subsection{History Guided Attention Module (HGAM)}

Adjacent frames in a video \tao{always} contain rich temporal information, as the positions of glass surfaces remain roughly consistent between neighboring frames. 
Effectively leveraging this temporal similarity can greatly enhance the robustness of \tao{our method}.
\tao{Therefore,} HGAM is proposed to enhance the current frame \tao{with} previous frames. 

Since the \tao{frame $I_N$} is acquired after the frames $I_{N-1}$ and $I_{N-2}$, its prediction $M_N$ \tao{can be improved with the help of the predictions for the previous frames, termed} $M_{N-2}$ and $M_{N-1}$. 
As shown in Fig.~\ref{fig:HGAM}, when the feature of current frame $G_i^N$ is used as the query ($Q_i^N \in \mathbb{R}^{H_iW_i \times C_i}$), the $K$ and $V$ are \tao{calculated} from $G_i^{N-2}$ and $G_i^{N-1}$, yielding rich features $\tilde{K}_i^N \in \mathbb{R}^{C_i \times H_iW_i}$ and $\tilde{V}_i^N \in \mathbb{R}^{H_iW_i \times C_i}$, \tao{as follows}:
\begin{align}
\tilde{K}_i^N &=[\mathrm{W_K}(G_i^{N-2})\odot \mathrm{W_K}({G_i^{N-1}}),\mathrm{W_K}(G_i^N)], \\
\tilde{V}_i^N &=[\mathrm{W_V}(G_i^{N-2})\odot \mathrm{W_V}({G_i^{N-1}}),\mathrm{W_V}(G_i^N)],
\end{align}
where $\odot$ denotes element-wise multiplication, and $[\cdot]$ represents channel-wise concatenation. \tao{These} features, \tao{along with the query $Q_i^N$}, are then fed into a self-attention to generate the temporal output feature $T_i^N$, \tao{as follows}:
\begin{equation}
    T_i^N = \mathrm{SelfAttn}(Q_i^N,\tilde{K}_i^N,\tilde{V}_i^N).
 \end{equation}

\subsection{Temporal Cross Attention Module (TCAM)}
%
TCAM employs a standard cross-attention mechanism to capture inter-frame dependencies, \tao{while minimizing redundant computation}. Specifically, $T_i^{N-1}=\text{TCAM}({G_i^{N-1}, G_i^{N-2})}$ and $T_i^{N-2}=\text{TCAM}({G_i^{N-2}, G_i^{N})}$. The cross-attention between frame $I_{N-1}$ and $I_{N-2}$ captures short-term temporal dependencies and motion trends, while the cross-attention between frame $I_{N-2}$ and $I_{N}$ facilitates long-range temporal consistency and provides complementary information for recovering occluded or degraded regions. Taking $T_i^{N-1}$ as an example, the feature $G_i^{N-1}$ serves as the Query and the feature $G_i^{N-2}$ serves as the Key and Value to form the attention map. Then, the \tao{cross-}attention map serves as a residual term to \tao{produce} the output features, \tao{as follows:}
\begin{equation}
T_i^{N-1} = G_i^{N-1} + \tao{\mathrm{CrossAttn}}(G_i^{N-1},G_i^{N-2},G_i^{N-2}).
\end{equation}
and the process for the $T_i^{N-2}$ is similar.

\begin{figure}[t]
\centerline{\includegraphics[width=0.45
\textwidth]{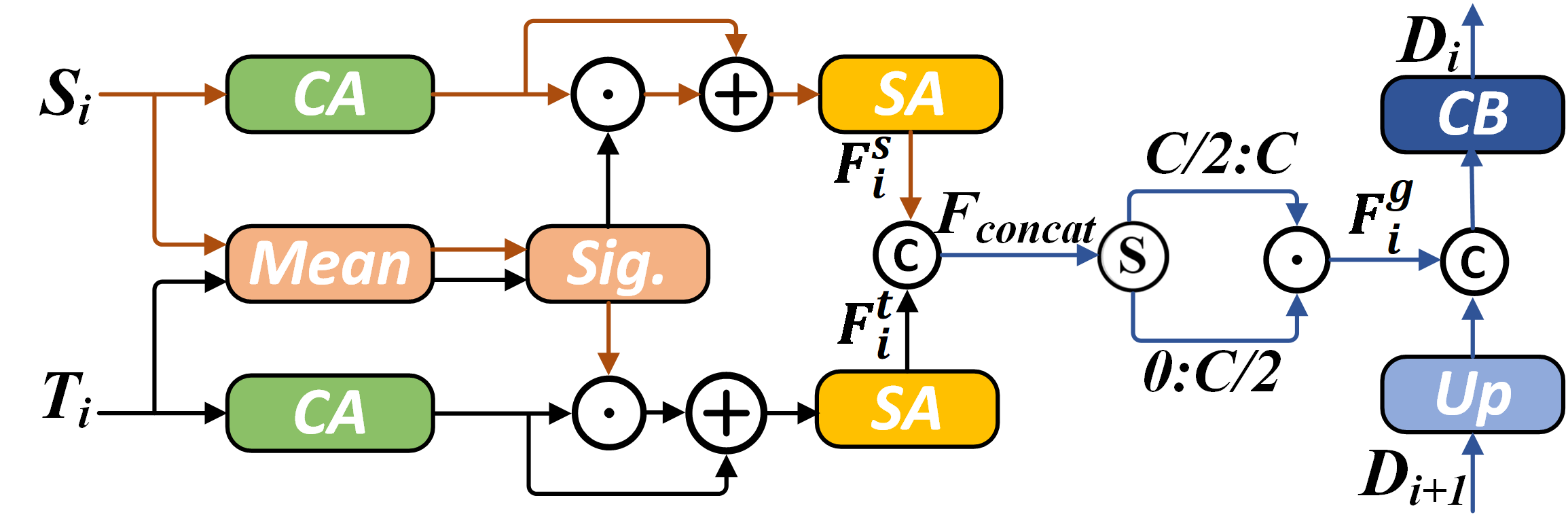}}
\caption{The structure of our TSD.}
\label{fig:TSD}
\end{figure}

\subsection{Temporal-Spatial Decoder (TSD)}
The temporal feature set obtained \tao{by} HGAM and TCAM has a channel configuration of ${\{C_i}\}_{i=1}^4={\{2^{i-1}C_1}\}_{i=1}^4$, while the spatial features produced by CMFM\tao{s} share a uniform channel dimension of $C_1$ across all scales. This channel inconsistency between the temporal and spatial domains hinders direct feature fusion. Thus, TSD is proposed to balance the channel dimensions of both domains in the process of feature fusion and decoding.

As shown in Fig.~\ref{fig:TSD}, the input temporal features $T_i \in \mathbb{R}^{H_i \times W_i \times C_i}$ and spatial features $S_i \in \mathbb{R}^{H_i \times W_i \times C_1}$ are fed into a Channel Attention module ($\text{CA}(\cdot)$) \tao{for enhancement.}
\tao{Then, feature weights are generated via a channel-wise mean and subsequent sigmoid activation. These weights then enhance the corresponding feature map through element-wise multiplication.}
%
This process can be formulated as:
\begin{gather}
F^t_i = \mathrm{SA}(\mathrm{CA}(T_i)\odot \mathrm{Sigmoid}(\mathrm{M}(S_i))+\mathrm{CA}(T_i)),\\
F^s_i = \mathrm{SA}(\mathrm{CA}(S_i)\odot \mathrm{Sigmoid}(\mathrm{M}(T_i))+\mathrm{CA}(S_i)).
\end{gather}
\tao{$\mathrm{SA}(\cdot)$ and $\mathrm{CA}(\cdot)$ represent Spatial Attention and Channel Attention, respectively. $\mathrm{M}(\cdot)$ denotes channel-wise mean function.}
The features $F^t_i$ and $F^s_i$ are \tao{then} concatenated along the channel dimension. 
However, these features are unbalanced as the temporal information still dominates the features. Inspired by the simple gate of the NAFNet~\cite{chen2022NAFBlock}, we evenly split the concatenated features into two halves along the channel dimension. Then, each corresponding pair of channels from the two halves is combined through element-wise multiplication to generate the gated output features $F^g_i$. This process can be formulated as:
\begin{align}
F_{concat}&= \mathrm{Concat}(F^t_i,F^s_i),\\
F_{i}^{g} &= F_{concat}^{[:C/2]}{\odot}F_{concat}^{[C/2:C]},
\end{align}
where $\odot$ denotes element-wise multiplication. 

The output from the previous decoding layer is upsampled and fused with the current gated output $F^g$, and then passed through a Convolution block with Batch Normalization, \tao{termed $\mathrm{CB}(\cdot)$}, to obtain the decoded feature for this layer. This process can be formulated as follows:
\begin{equation}
    D_i =     
    \begin{cases}
        F_4^g  &i=4 \\
        \mathrm{CB}(\mathrm{Concat}(F_i^g, \mathrm{UP}(D_{i+1}))) &i=1,2,3.
    \end{cases}
\end{equation}
Finally, we apply a 1×1 convolution to the feature $D_1$ to generate the segmentation mask of the glass region.
\subsection{Loss Function}
We employ BCE loss~\cite{de2005CE} and IoU loss~\cite{qin2019BASNet} to supervise the predicted results at the pixel level and the region level, respectively. The overall loss consists of the loss for the primary glass surface mask $P_{N-1}$ and the losses for the predicted masks of three adjacent frames. 

\taore{The} loss for the primary mask \tao{$P_{N-1}$} is defined as:
\begin{equation}
    \mathcal{L}_{P} = \mathcal{L}_{BCE}(P_{N-1}, \hat{M}_{N-1})+\mathcal{L}_{IoU}(P_{N-1},\hat{M}_{N-1}),
\end{equation}
where $\hat{M}_{N-1}$ denotes the ground truth of the glass surface mask in the frame of N-1.

\tao{We also} employ the same loss for the prediction across the three adjacent frames. This can be formulated as:
\begin{equation}
    \mathcal{L}_{M} = \sum_{i=N-2}^{N}(\mathcal{L}_{\text{BCE}}(M_i, \hat{M}_i) + \mathcal{L}_{\text{IoU}}(M_i, \hat{M}_i)),
\end{equation}
where $M$ denotes the prediction of the glass surface mask.

The total loss of the network consists of the above two losses and can be defined as:
\begin{equation}
    \mathcal{L} = \alpha \mathcal{L}_P + \mathcal{L}_M,
\end{equation}
where $\alpha$ is \tao{empirically} set to $1/8$ \tao{for balancing} the weights between $\mathcal{L}_P$ and $\mathcal{L}_M$.

\newcommand{\newsubheight}{0.118}
\begin{figure}[b]
    \renewcommand{\tabcolsep}{0.8pt}
    \renewcommand\arraystretch{0.6}
    \centering
        \begin{tabular}{cccccccc}
                \includegraphics[height=\newsubheight\linewidth, keepaspectratio]{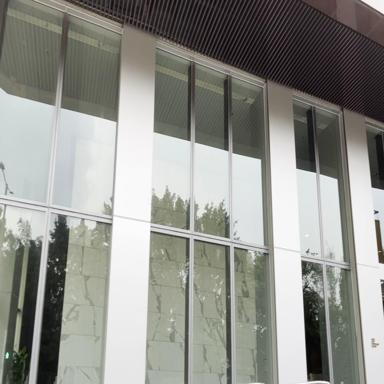}&
                \includegraphics[height=\newsubheight\linewidth, keepaspectratio]{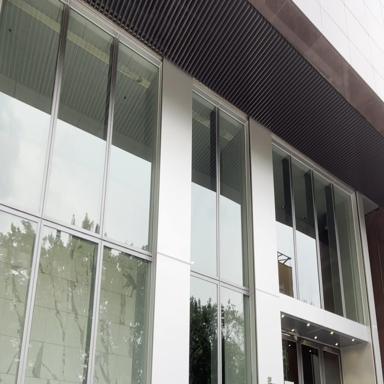}&
                \hspace{1.0pt}
                \includegraphics[height=\newsubheight\linewidth, keepaspectratio]{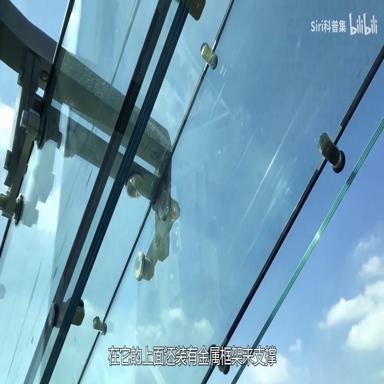}&
                \includegraphics[height=\newsubheight\linewidth, keepaspectratio]{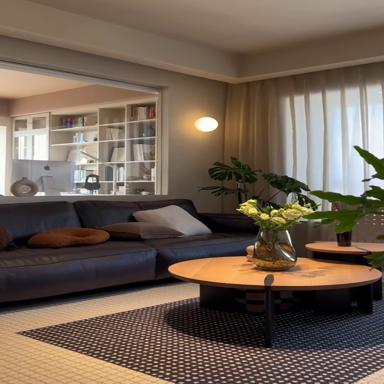}&
                \includegraphics[height=\newsubheight\linewidth, keepaspectratio]{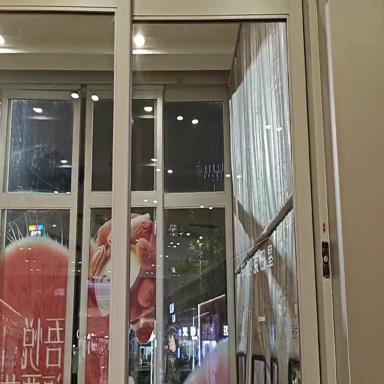}&
                \includegraphics[height=\newsubheight\linewidth, keepaspectratio]{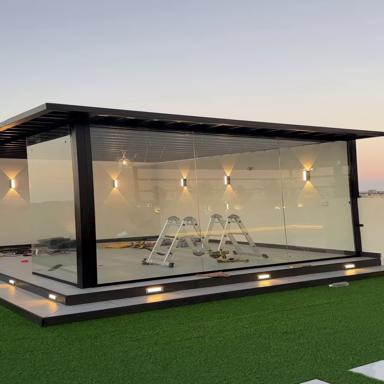}&
                \includegraphics[height=\newsubheight\linewidth, keepaspectratio]{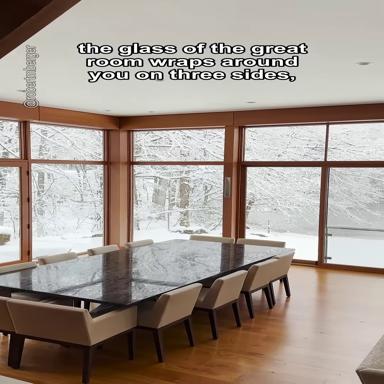}&
                \includegraphics[height=\newsubheight\linewidth, keepaspectratio]{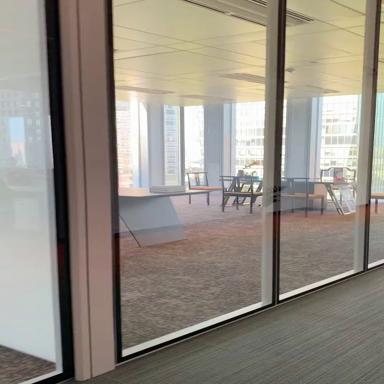}
                \\
  
                \includegraphics[height=\newsubheight\linewidth, keepaspectratio]{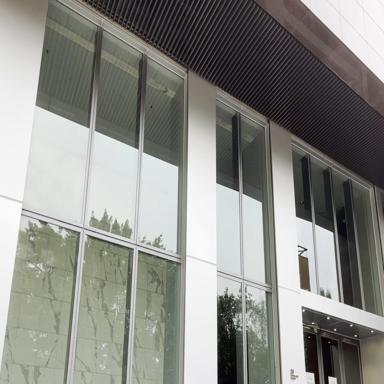}&
                \includegraphics[height=\newsubheight\linewidth, keepaspectratio]{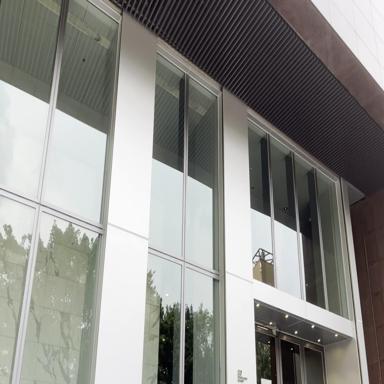}&
                \hspace{1.0pt}
                \includegraphics[height=\newsubheight\linewidth, keepaspectratio]{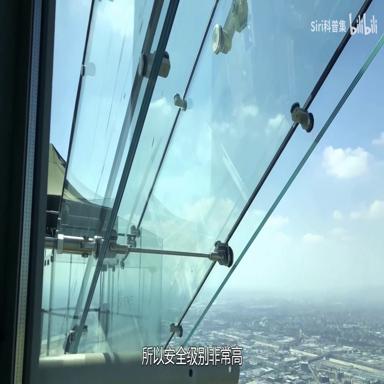}&
                \includegraphics[height=\newsubheight\linewidth, keepaspectratio]{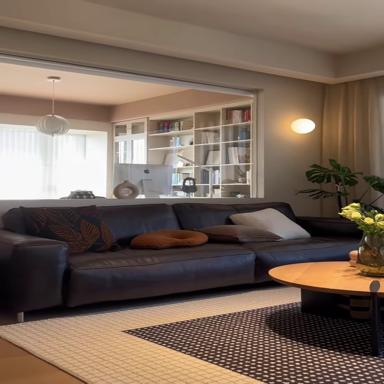}&
                \includegraphics[height=\newsubheight\linewidth, keepaspectratio]{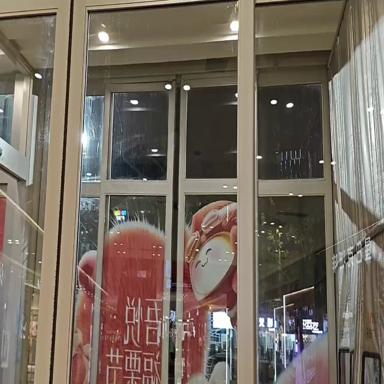}&
                \includegraphics[height=\newsubheight\linewidth, keepaspectratio]{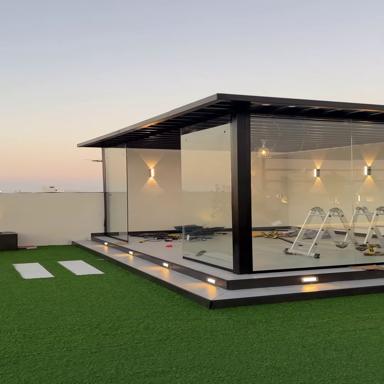}&
                \includegraphics[height=\newsubheight\linewidth, keepaspectratio]{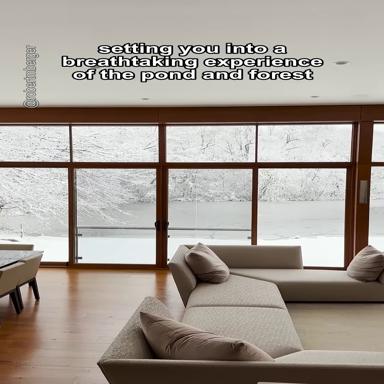}&
                \includegraphics[height=\newsubheight\linewidth, keepaspectratio]{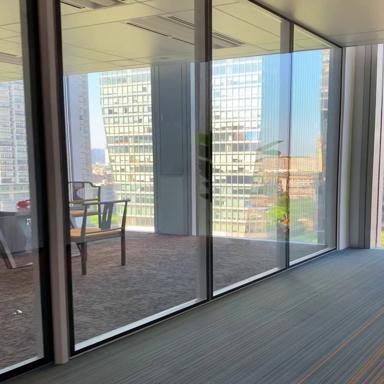}
                \\ 

                \includegraphics[height=\newsubheight\linewidth, keepaspectratio]{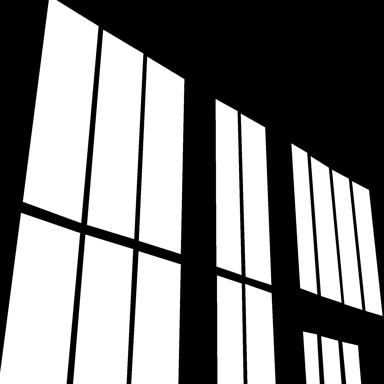}&
                \includegraphics[height=\newsubheight\linewidth, keepaspectratio]{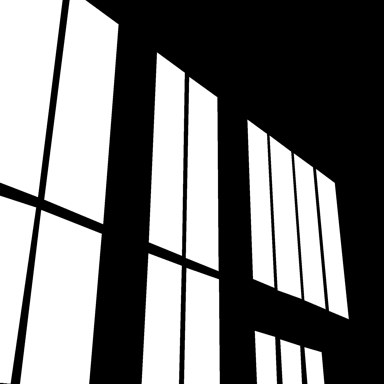}&
                \hspace{1.0pt}
                \includegraphics[height=\newsubheight\linewidth, keepaspectratio]{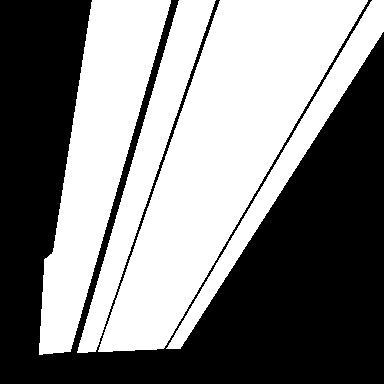}&
                \includegraphics[height=\newsubheight\linewidth, keepaspectratio]{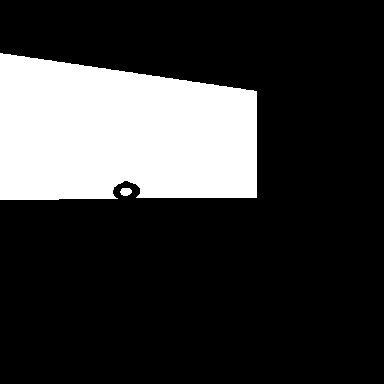}&
                \includegraphics[height=\newsubheight\linewidth, keepaspectratio]{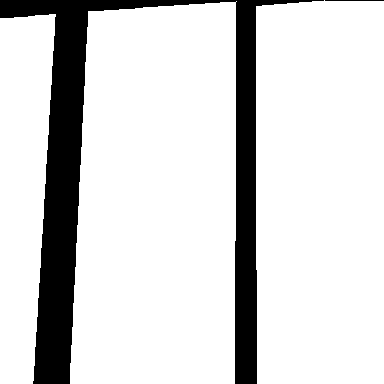}&
                \includegraphics[height=\newsubheight\linewidth, keepaspectratio]{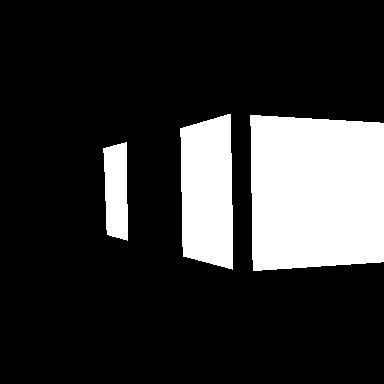}&
                \includegraphics[height=\newsubheight\linewidth, keepaspectratio]{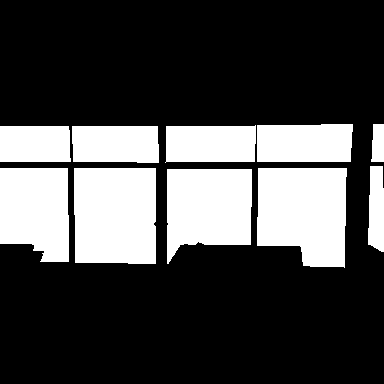}&
                \includegraphics[height=\newsubheight\linewidth, keepaspectratio]{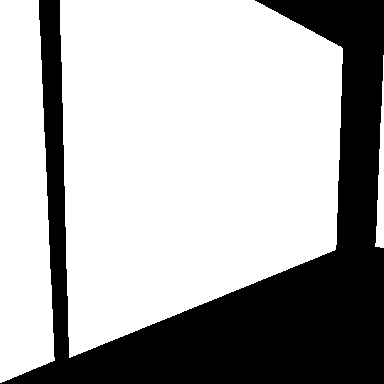}
                \\

                \multicolumn{2}{c}{\small{VGSD-D}}&
                \multicolumn{6}{c}{\small{MVGD-D (Ours)}}\\
                
        \end{tabular}
    \caption{
    \lyw{Comparison between VGSD-D($1$st-$2$nd columns) and our MVGD-D($3$rd-$8$th columns). For each scene, from top to bottom, we show two selected frames and the mask for the second frame.}
    }
    \label{fig:examples_from_dataset}
\end{figure}

\begin{table}[ht]
\centering
\small
\renewcommand\arraystretch{1.05}
\setlength{\tabcolsep}{3.6pt} 
\begin{tabular}{l l l c c}
\toprule
Task & Dataset & Publication & Videos & Frames \\
\midrule
\multirow{2}{*}{VGSD} & VGSD-D  & AAAI'24 & 297 & 19166 \\
                      & PVG-117 & ICCV'23 & 117 & 21485 \\
\midrule
\multirow{2}{*}{VMD}  & MMD      & CVPR'24 & 37  & 9727 \\
                      & VMD-D   & CVPR'23 & 269 & 15066 \\
\midrule
VGSD & MVGD-D(Ours)    & - & 312 & 19268 \\
\bottomrule
\end{tabular}
\caption{Comparison of Datasets for VGSD\tao{/VMD} task.}
\label{tab:dataset_comparison}
\end{table}

\renewcommand{\newsubwidth}{0.47}
\begin{figure}[htbp]
	\renewcommand{\tabcolsep}{1.0pt}
	\renewcommand\arraystretch{0.6}
    \centering
            \begin{tabular}{cc}
                \includegraphics[width=0.43\linewidth]{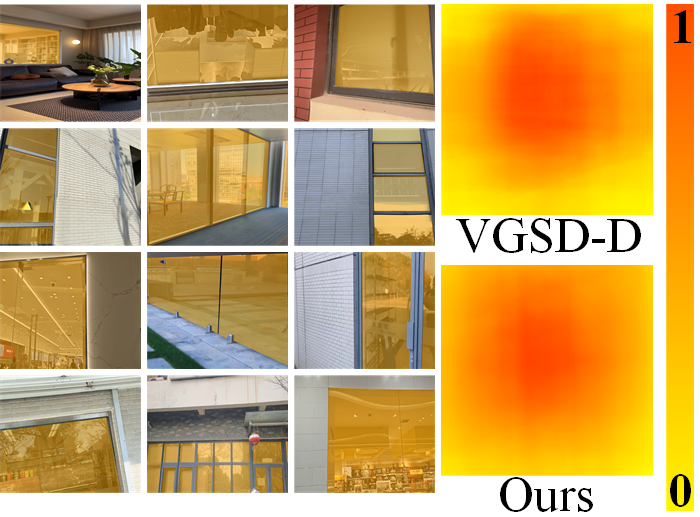}
                &
                \includegraphics[width=0.5\linewidth]{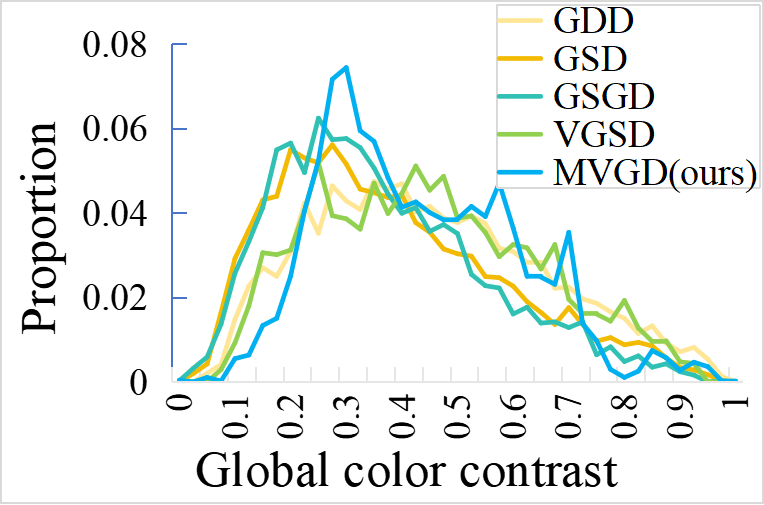}
                \\
                \fontsize{9.0pt}{\baselineskip}\selectfont{(a) Glass Region Distribution}&
                \fontsize{9.0pt}{\baselineskip}\selectfont{(b) Color Contrast Distribution}
            \end{tabular}
\caption{Statistics of our MVGD-D.}
\label{fig:dataset_statistics}
\end{figure}

\begin{figure*}[htbp]
    \renewcommand{\tabcolsep}{0.8pt}
    \renewcommand\arraystretch{0.6}
    \renewcommand{\newsubwidth}{0.078}
    \centering
    \begin{tabular}{cccccccccccc}
        \multirow{6}{*}[+9ex]{\includegraphics[width=0.0585\textwidth]{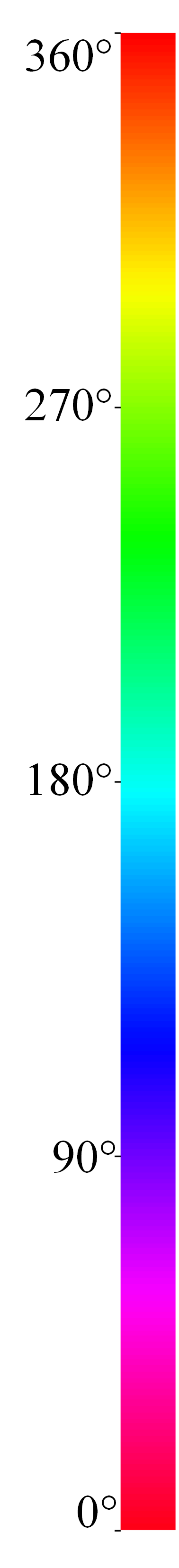}}&
        \includegraphics[width=\newsubwidth\linewidth]{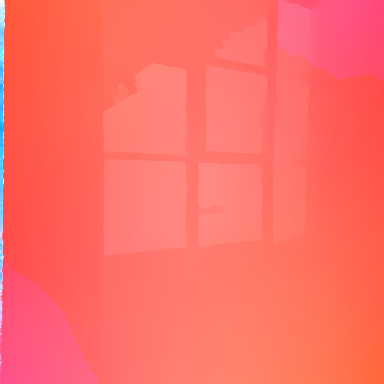}&
        \includegraphics[width=\newsubwidth\linewidth]{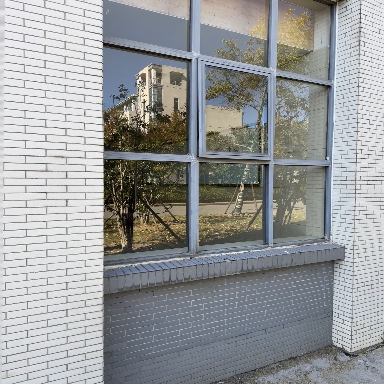}&
        \includegraphics[width=\newsubwidth\linewidth]{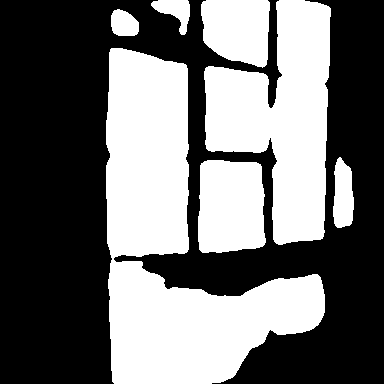}&
        \includegraphics[width=\newsubwidth\linewidth]{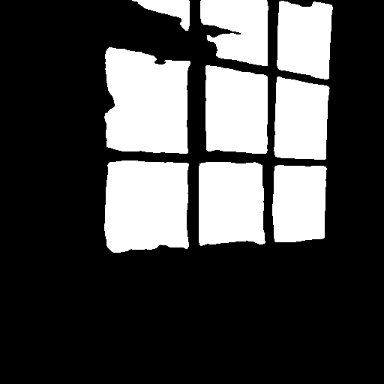}&
        \includegraphics[width=\newsubwidth\linewidth]{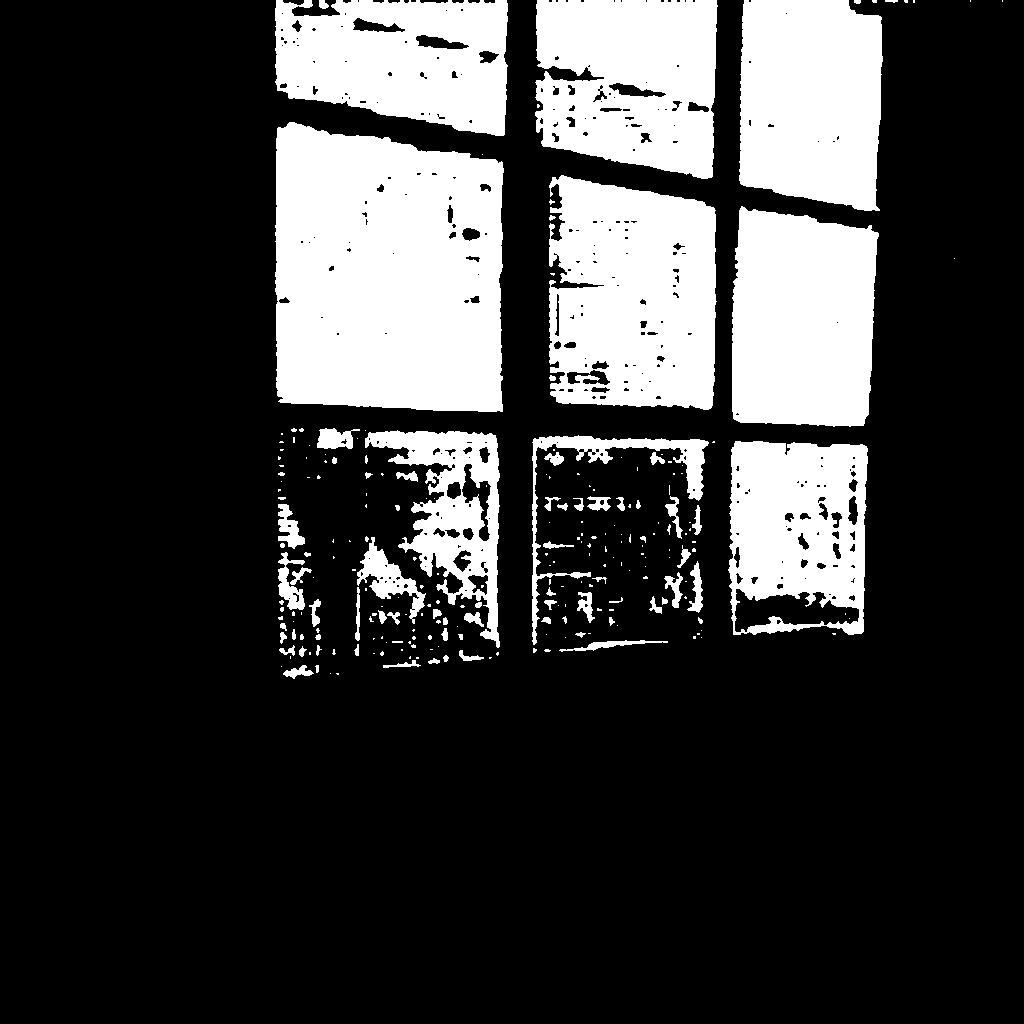}&
        \includegraphics[width=\newsubwidth\linewidth]{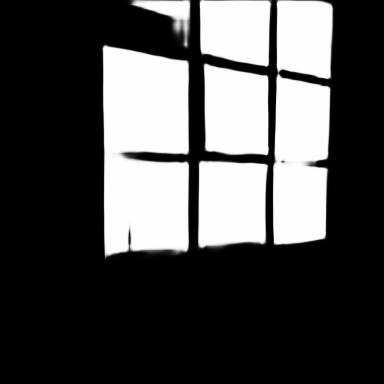}&
        \includegraphics[width=\newsubwidth\linewidth]{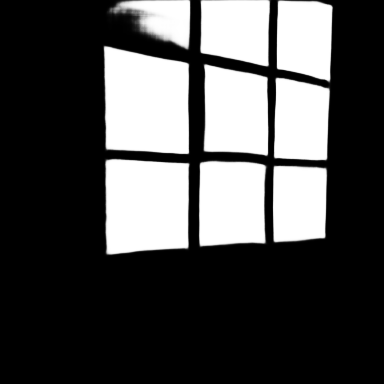}&
        \includegraphics[width=\newsubwidth\linewidth]{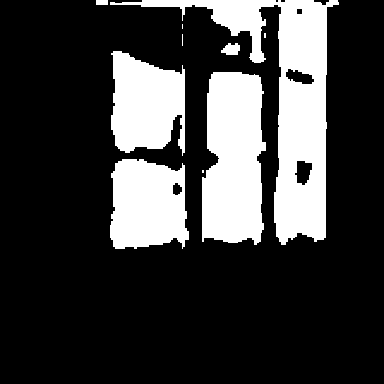}&
        \includegraphics[width=\newsubwidth\linewidth]{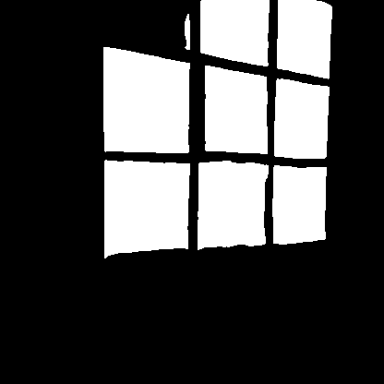}&
        \includegraphics[width=\newsubwidth\linewidth]{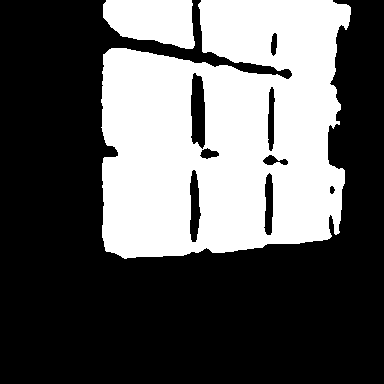}&
        \includegraphics[width=\newsubwidth\linewidth]{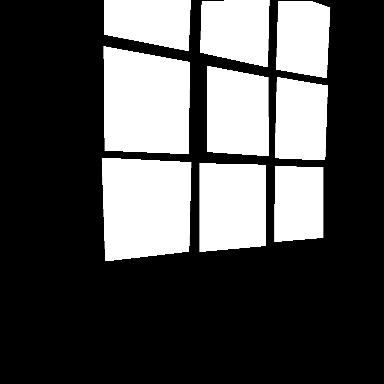}
        \\
        &\includegraphics[width=\newsubwidth\linewidth]{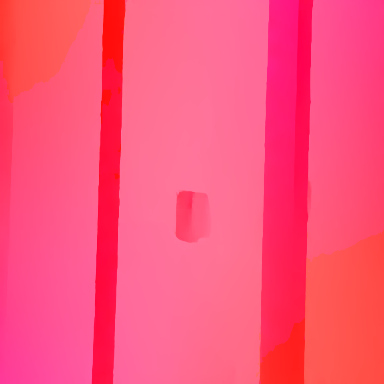}&
        \includegraphics[width=\newsubwidth\linewidth]{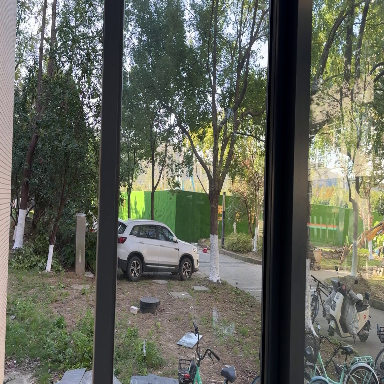}&
        \includegraphics[width=\newsubwidth\linewidth]{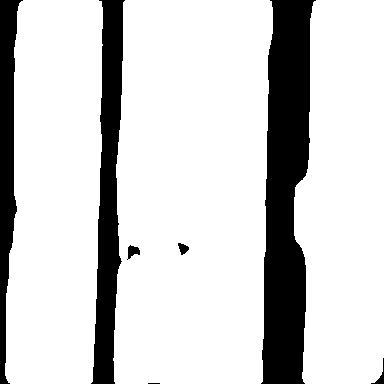}&
        \includegraphics[width=\newsubwidth\linewidth]{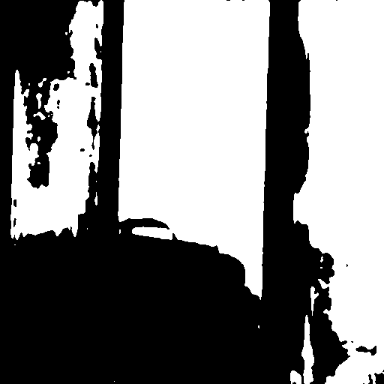}&
        \includegraphics[width=\newsubwidth\linewidth]{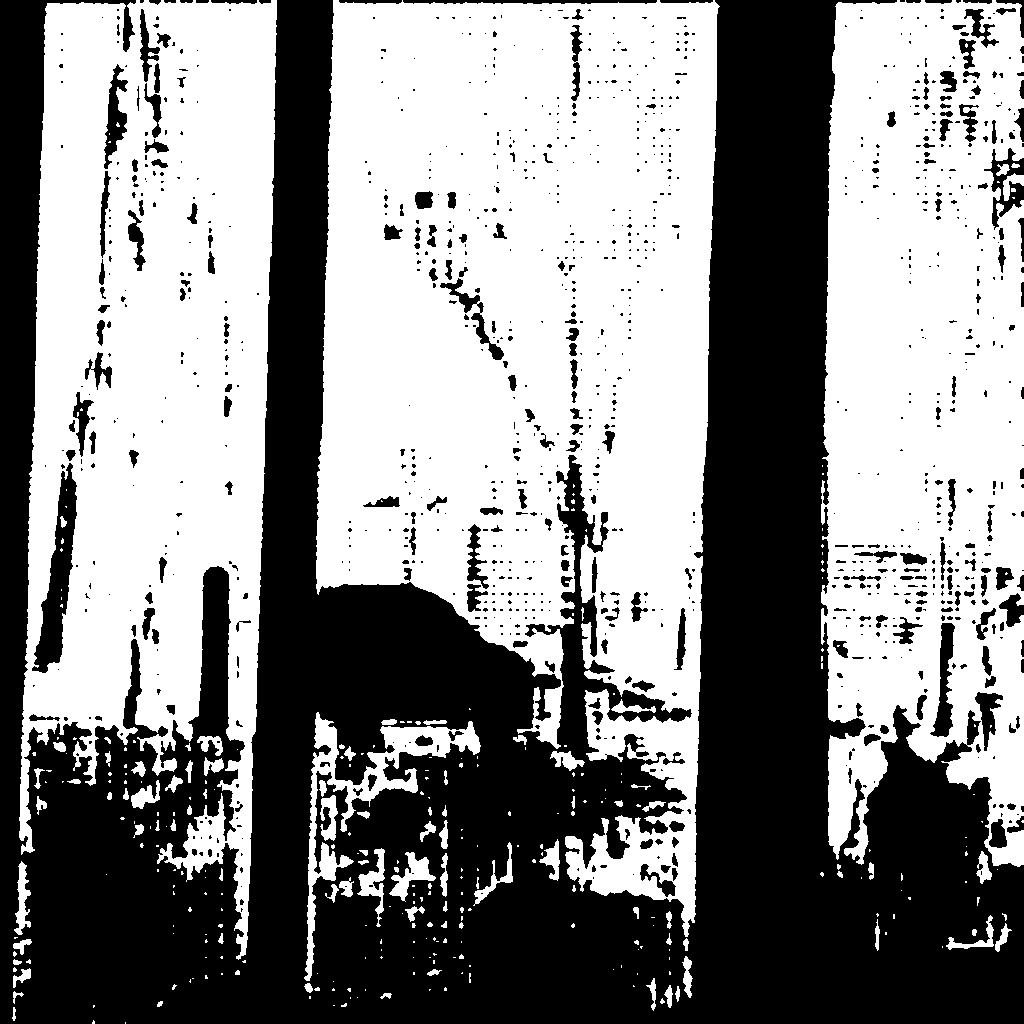}&
        \includegraphics[width=\newsubwidth\linewidth]{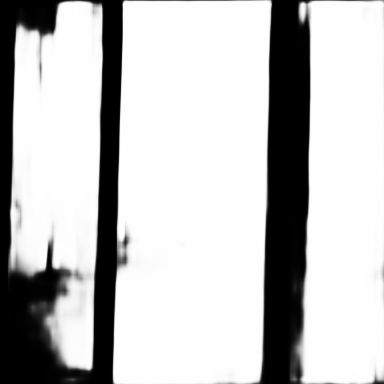}&
        \includegraphics[width=\newsubwidth\linewidth]{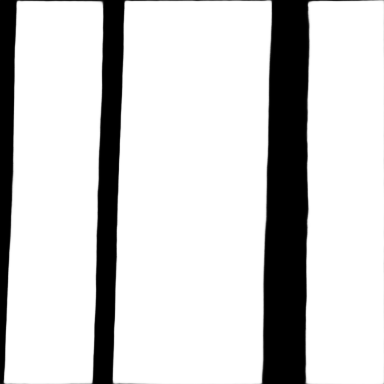}&
        \includegraphics[width=\newsubwidth\linewidth]{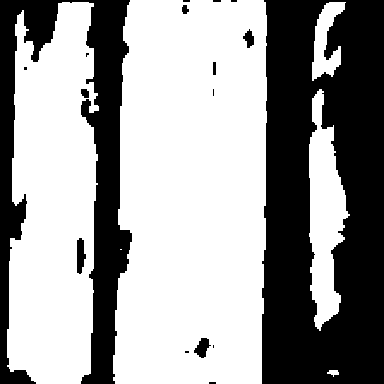}&
        \includegraphics[width=\newsubwidth\linewidth]{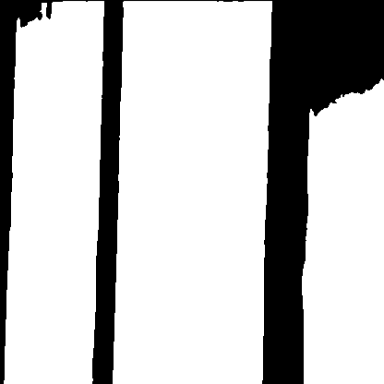}&
        \includegraphics[width=\newsubwidth\linewidth]{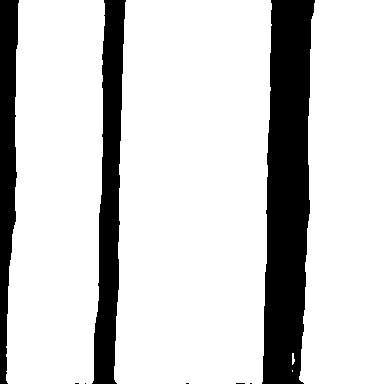}&
        \includegraphics[width=\newsubwidth\linewidth]{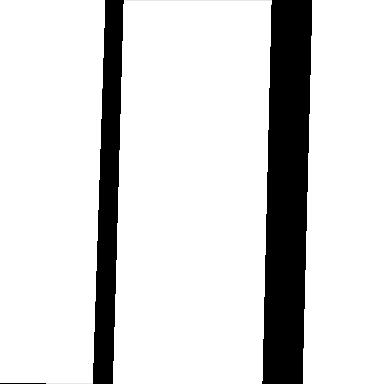}
        \\
        &\includegraphics[width=\newsubwidth\linewidth]{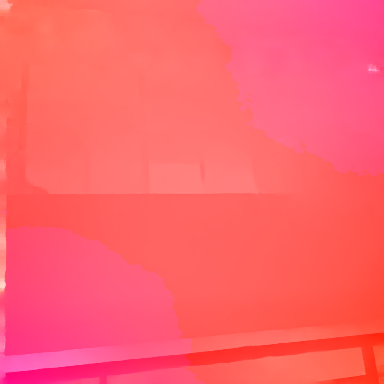}&
        \includegraphics[width=\newsubwidth\linewidth]{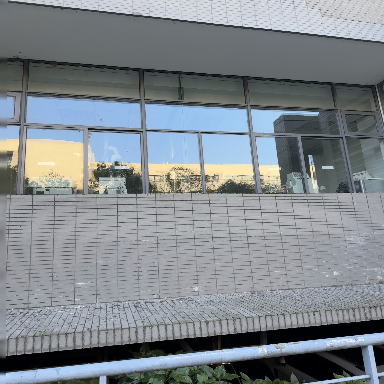}&
        \includegraphics[width=\newsubwidth\linewidth]{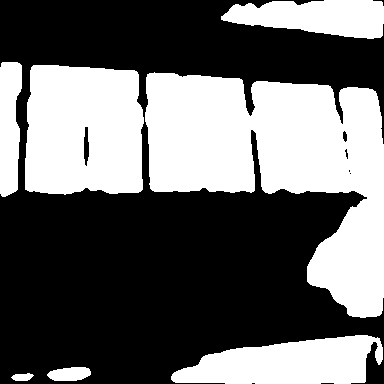}&
        \includegraphics[width=\newsubwidth\linewidth]{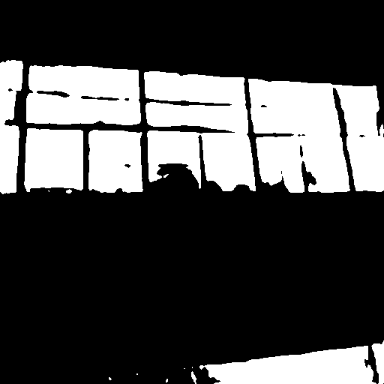}&
        \includegraphics[width=\newsubwidth\linewidth]{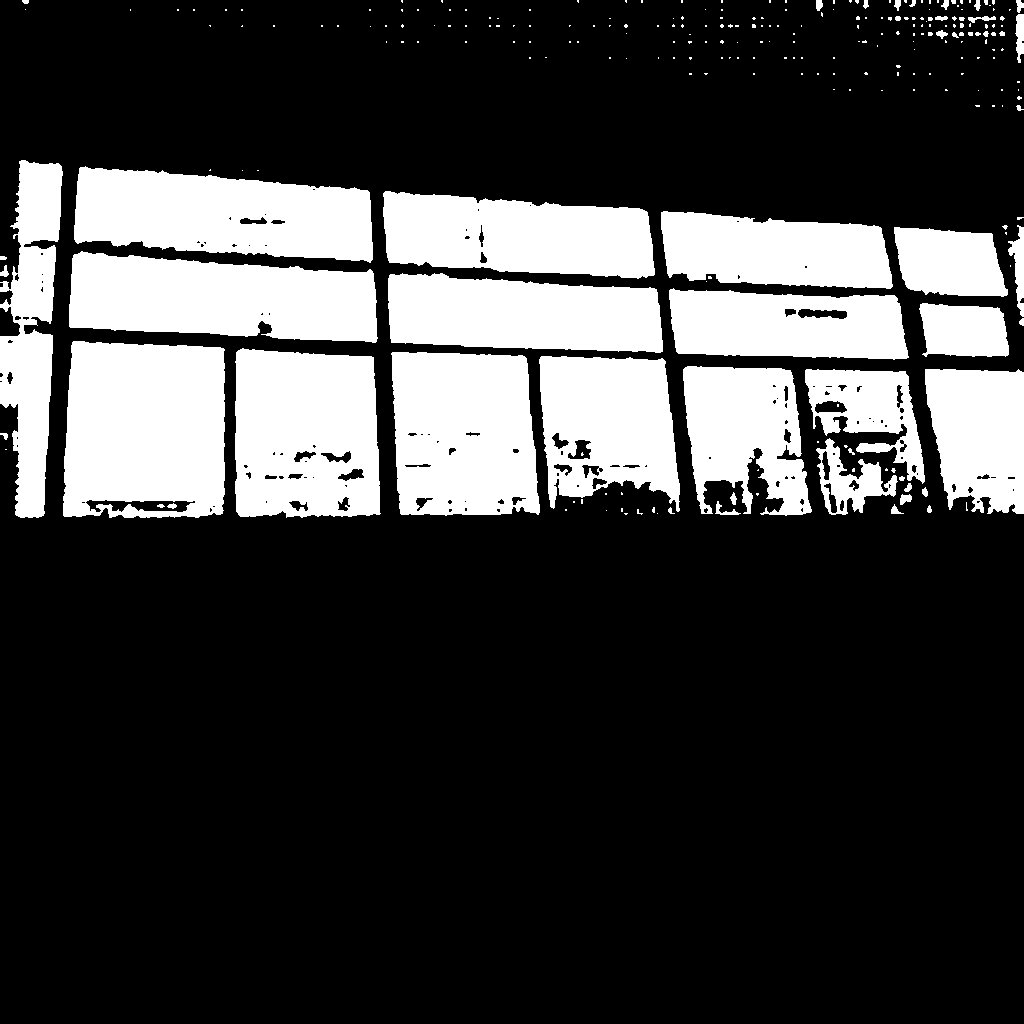}&
        \includegraphics[width=\newsubwidth\linewidth]{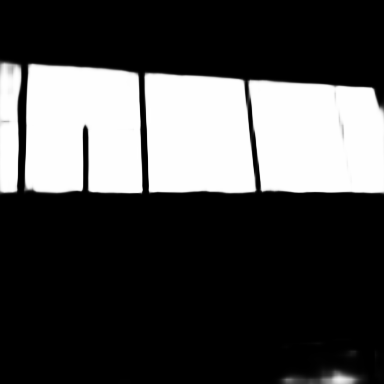}&
        \includegraphics[width=\newsubwidth\linewidth]{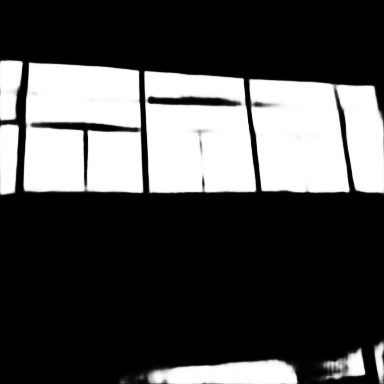}&
        \includegraphics[width=\newsubwidth\linewidth]{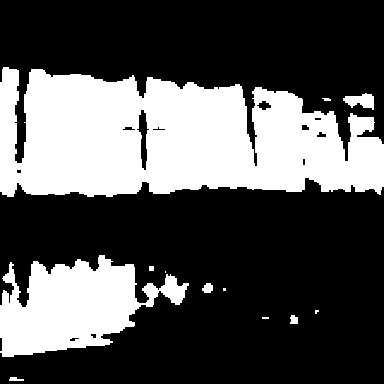}&
        \includegraphics[width=\newsubwidth\linewidth]{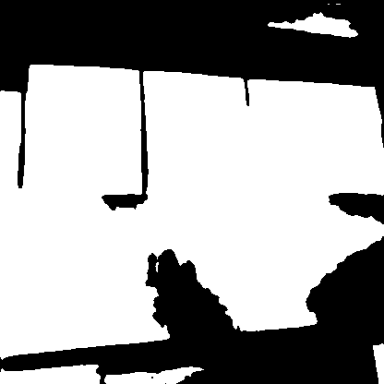}&
        \includegraphics[width=\newsubwidth\linewidth]{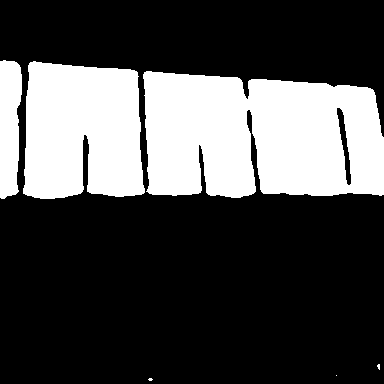}&
        \includegraphics[width=\newsubwidth\linewidth]{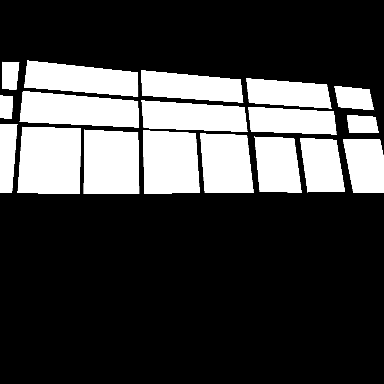}
        \\
        &\includegraphics[width=\newsubwidth\linewidth]{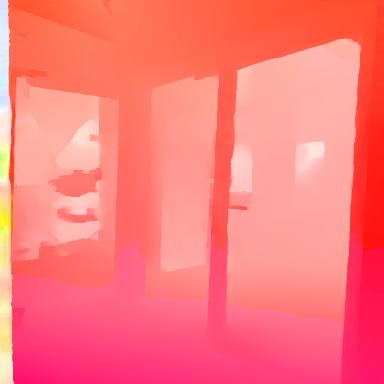}&
        \includegraphics[width=\newsubwidth\linewidth]{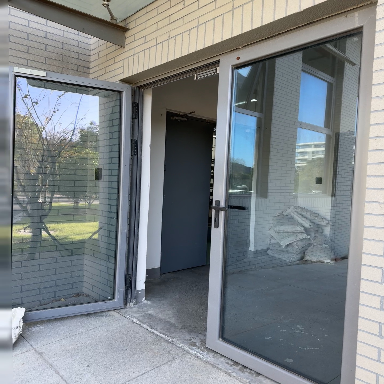}&
        \includegraphics[width=\newsubwidth\linewidth]{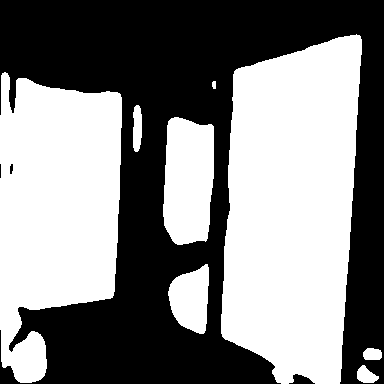}&
        \includegraphics[width=\newsubwidth\linewidth]{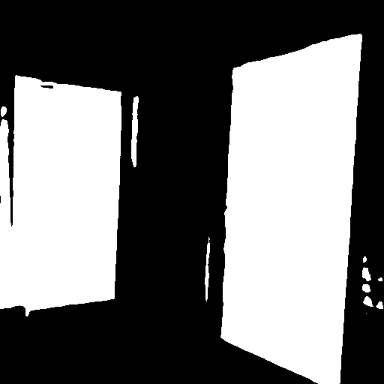}&
        \includegraphics[width=\newsubwidth\linewidth]{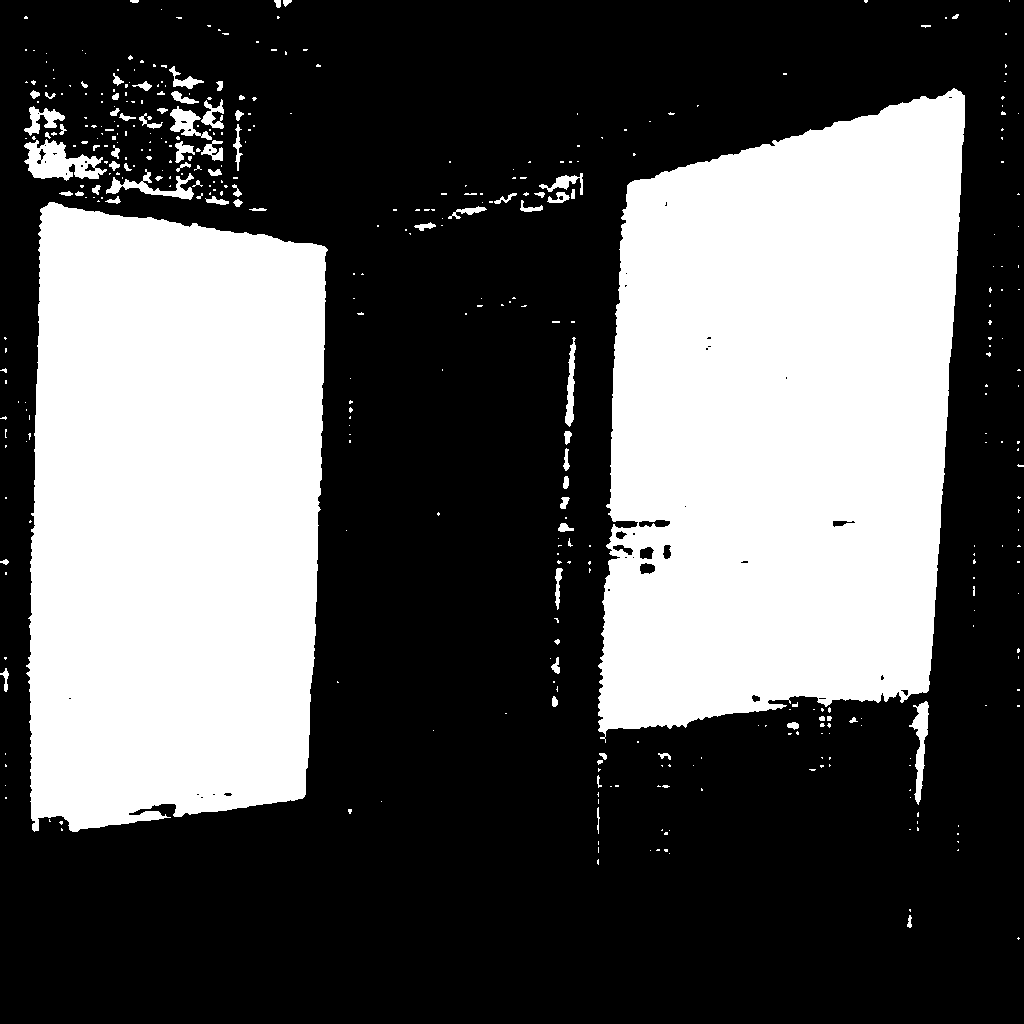}&
        \includegraphics[width=\newsubwidth\linewidth]{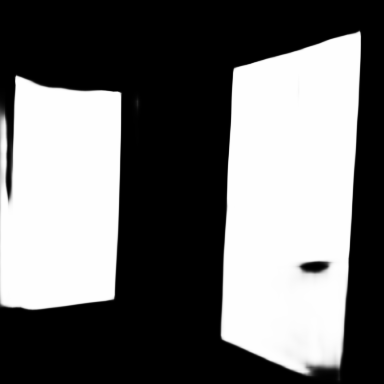}&
        \includegraphics[width=\newsubwidth\linewidth]{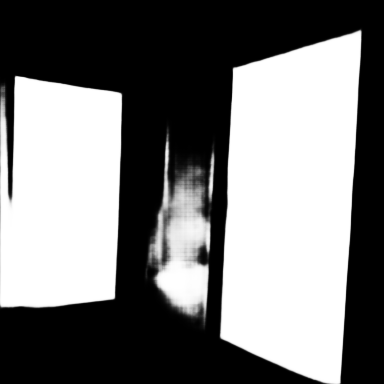}&
        \includegraphics[width=\newsubwidth\linewidth]{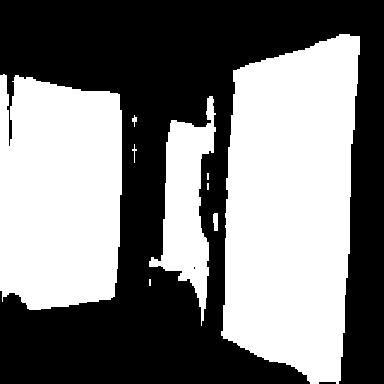}&
        \includegraphics[width=\newsubwidth\linewidth]{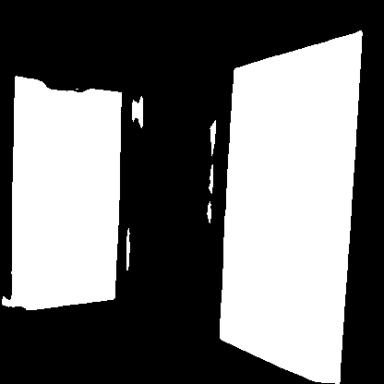}&
        \includegraphics[width=\newsubwidth\linewidth]{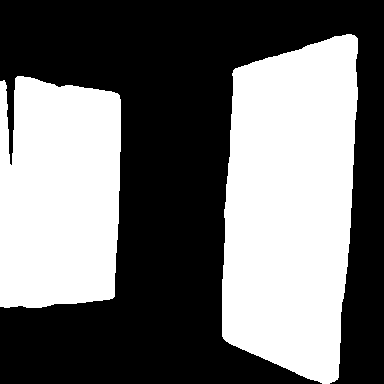}&
        \includegraphics[width=\newsubwidth\linewidth]{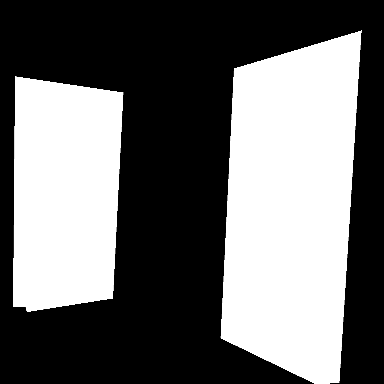}
        \\
        &\includegraphics[width=\newsubwidth\linewidth]{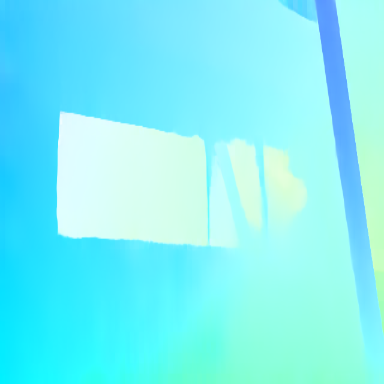}&
        \includegraphics[width=\newsubwidth\linewidth]{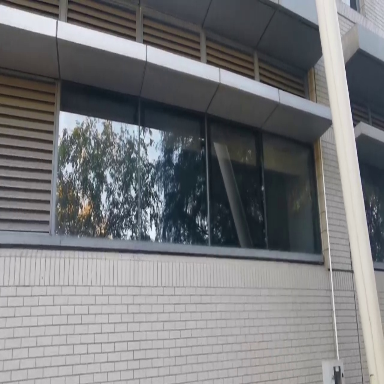}&
        \includegraphics[width=\newsubwidth\linewidth]{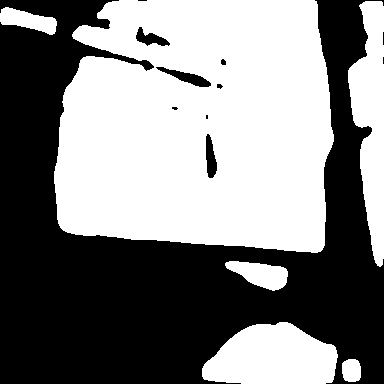}&
        \includegraphics[width=\newsubwidth\linewidth]{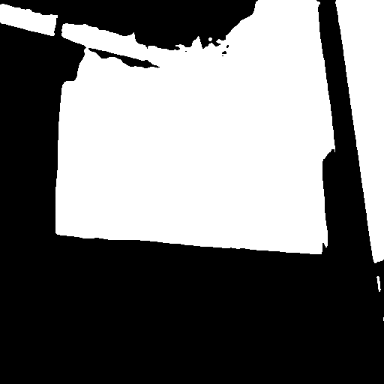}&
        \includegraphics[width=\newsubwidth\linewidth]{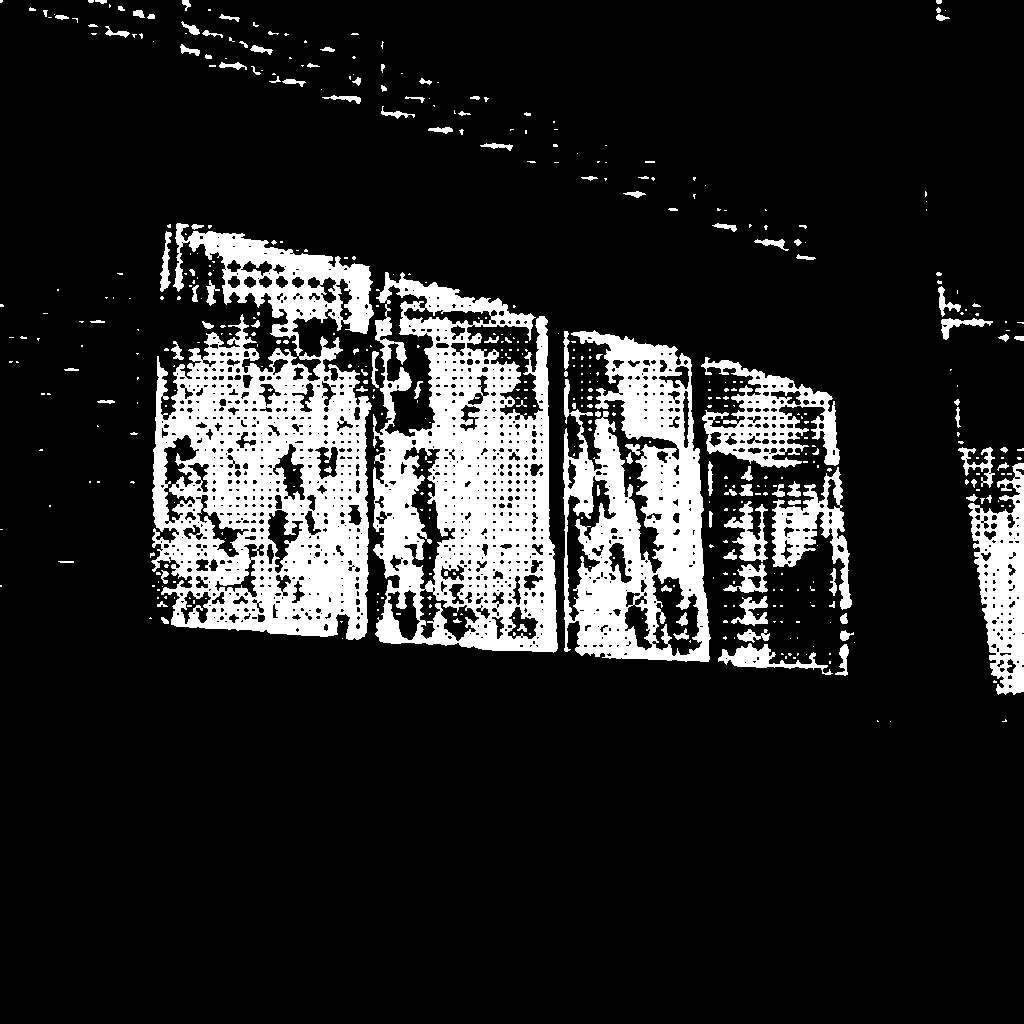}&
        \includegraphics[width=\newsubwidth\linewidth]{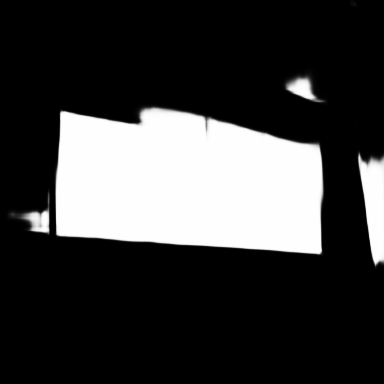}&
        \includegraphics[width=\newsubwidth\linewidth]{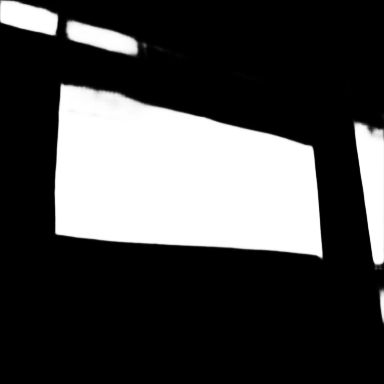}&
        \includegraphics[width=\newsubwidth\linewidth]{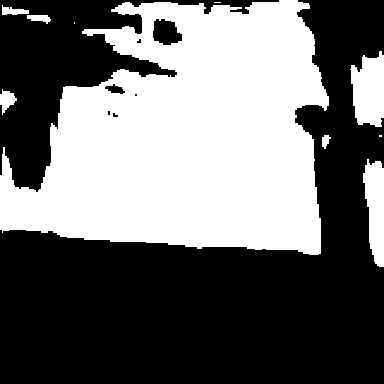}&
        \includegraphics[width=\newsubwidth\linewidth]{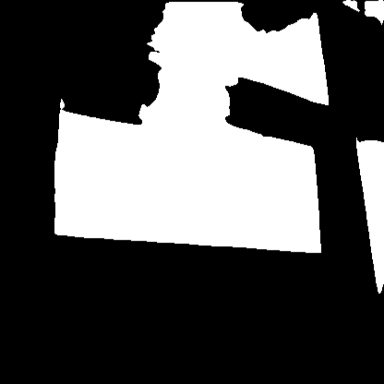}&
        \includegraphics[width=\newsubwidth\linewidth]{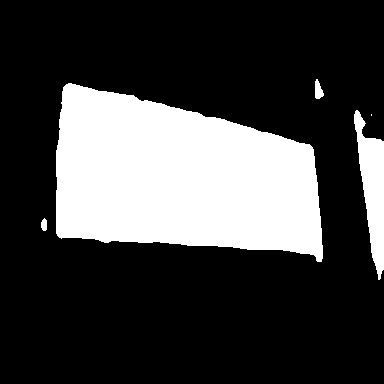}&
        \includegraphics[width=\newsubwidth\linewidth]{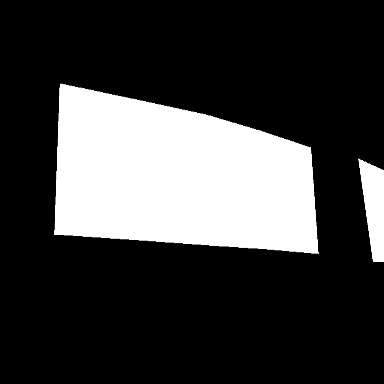}
        \\
        &\includegraphics[width=\newsubwidth\linewidth]{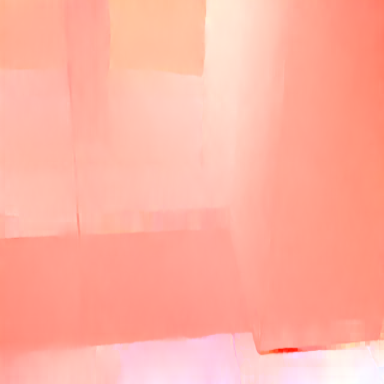}&
        \includegraphics[width=\newsubwidth\linewidth]{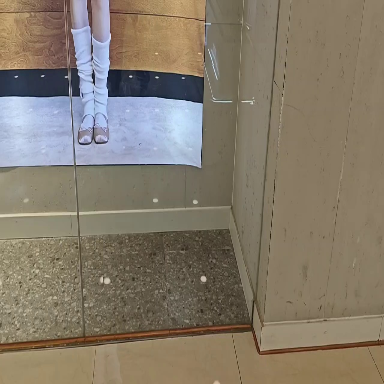}&
        \includegraphics[width=\newsubwidth\linewidth]{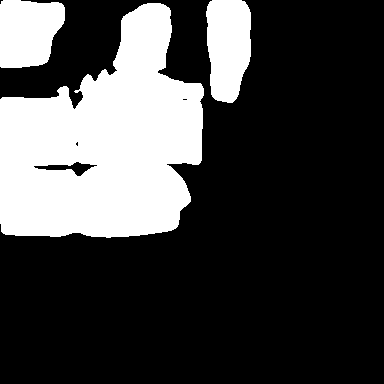}&
        \includegraphics[width=\newsubwidth\linewidth]{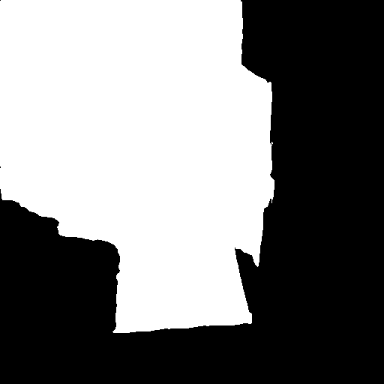}&
        \includegraphics[width=\newsubwidth\linewidth]{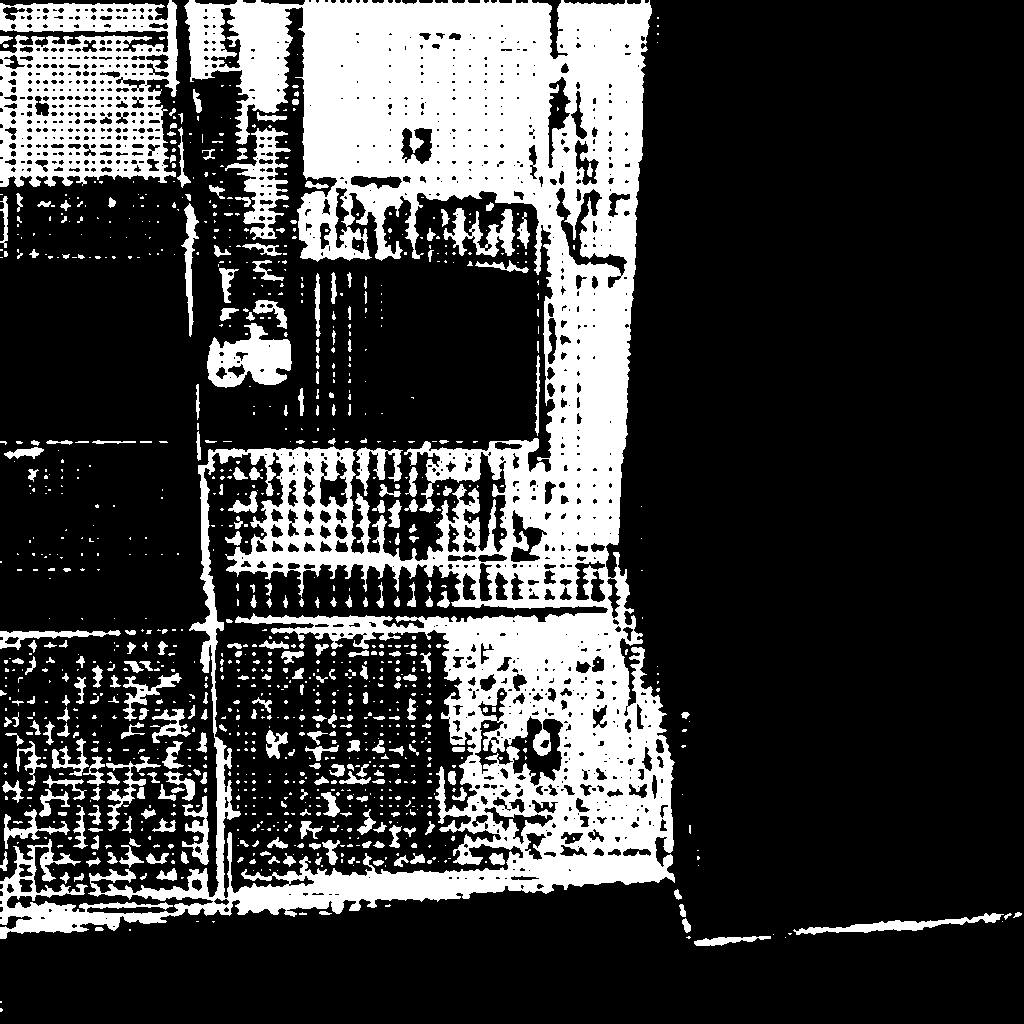}&
        \includegraphics[width=\newsubwidth\linewidth]{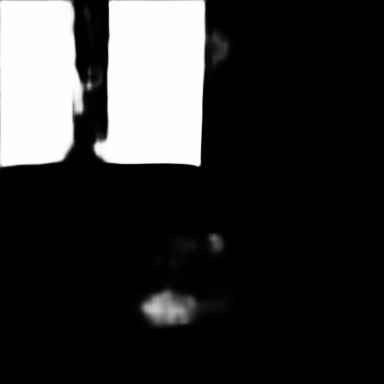}&
        \includegraphics[width=\newsubwidth\linewidth]{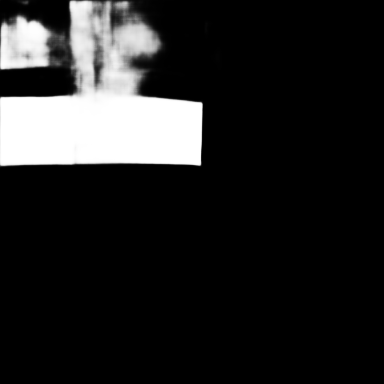}&
        \includegraphics[width=\newsubwidth\linewidth]{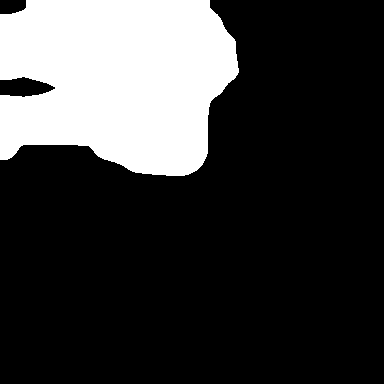}&
        \includegraphics[width=\newsubwidth\linewidth]{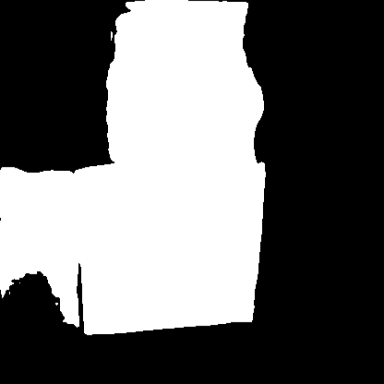}&
        \includegraphics[width=\newsubwidth\linewidth]{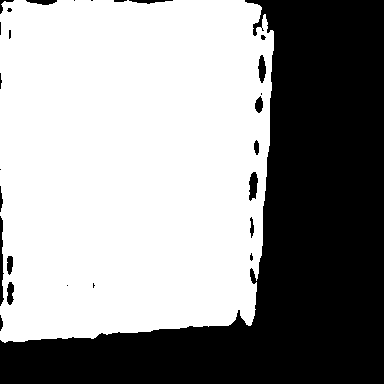}&
        \includegraphics[width=\newsubwidth\linewidth]{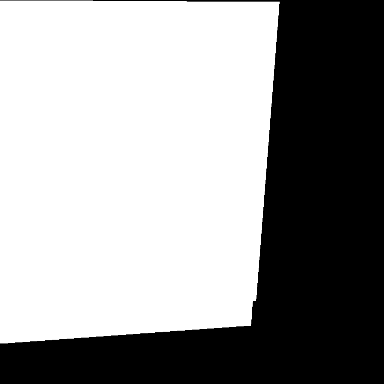}
        \\
        
        &\small{Flow Map}&
        \small{Frames}&
        \small{MINet}&
        \small{Isomer}&
        \small{SAM2}&
        \small{RFENet}&
        \small{Ghosting}&
        \small{MGVMD}&
        \small{VGSD}&
        \small{Ours}&
        \small{GT}\\
    \end{tabular}
    \caption{Qualitative comparison \tao{of our method and competing} methods on MVGD-D. The leftmost \tao{subfig} shows the optical flow color bar.}
    \label{fig:qualitative_comparison_with_colorbar_MVGSD}
\end{figure*}

\section{Dataset}
Though Liu \etal~\cite{liu2024VGSDNet} proposed \tao{the dataset} VGSD-D for VGSD, this dataset covers only a limited variety of scenes, as illustrated in Fig.~\ref{fig:examples_from_dataset}. 
Moreover, VGSD-D suffers from biased distributions of glass surface locations and overly pronounced color contrast between glass and non-glass regions. 
To alleviate these limitations, we construct a new large-scale dataset, named MVGD-D, which \tao{comprises $312$ real-world video clips ($19,268$ images)} with glass surfaces, manually annotated masks and flow maps produced by RAFT~\cite{teed2020raft}. \tao{MVGD-D covers} diverse \tao{static} scenes with dynamic \tao{camera} movements.

\subsection{Dataset Construction}

Our MVGD-D consists of $236$ real-world videos captured by handheld devices and $76$ videos sampled from streaming platforms (which are under the Creative Commons licenses). The handheld videos are recorded using DSLR cameras and smartphones, while the streaming videos are collected from platforms such as YouTube and Bilibili. Each video \tao{has a duration of} approximately $3\mbox{-}10$ seconds.
\taore{Videos were captured at $30$ FPS but sampled at $15$ FPS to balance redundancy and motion observability.}
Subsequently, experienced volunteers were engaged to carefully annotate video frames. 
The size of our dataset is comparable to that of \tao{existing representative} video glass datasets. A comparison of the VGSD datasets (VGSD-D~\cite{liu2024VGSDNet} and PVG117~\cite{Qiao2023PVGNet}) and the VMD \tao{datasets} (MMD~\cite{warren2024MGVMD} and VMD-D~\cite{lin2023VMDNet}) is shown in \tao{Tab.}~\ref {tab:dataset_comparison}.

\subsection{Dataset Analysis}
\begin{itemize}
\item \textbf{Glass Location.} As shown in Fig.~\ref{fig:dataset_statistics} (a), MVGD-D exhibits diverse glass surface locations, covering nine distinct spatial patterns. We overlap all masks to obtain the binary distribution of all \tao{glass/mirror} surfaces \tao{within} VGSD-D and MVGD-D, respectively. The distribution \tao{exhibited} in Fig.~\ref{fig:dataset_statistics} (a) shows that though both datasets exhibit glass primarily in the center region, the more uniform color \tao{distribution } in MVGD-D \tao{reflects} a more balanced \tao{underlying} distribution, which suggests that MVGD-D is more effective in avoiding the center bias problem.

\item \textbf{Color Contrast.} A lower color contrast between glass and non-glass regions indicates a higher similarity between the glass surface and its surrounding scene, thereby increasing the difficulty for VGSD. We compute the $\chi^2$ distance to evaluate the color contrast across several glass surface datasets(GDD~\cite{mei2020GDNet}, GSD~\cite{lin2021GSDNet}, GSGD~\cite{yan2025GhostingNet}, VGSD~\cite{liu2024VGSDNet}). Fig.~\ref{fig:dataset_statistics} (b) \tao{shows that} the contrast values in our dataset are mainly distributed \tao{within} the range of $[0.2, 0.6]$, suggesting that MVGD-D is more varied and challenging for VGSD.
\end{itemize}

\section{Experiments}
\subsection{Implementations}
\tao{All methods were retrained and evaluated on the two VGSD datasets using PyTorch on an NVIDIA RTX 4090 (24GB).}
\tao{We resized all frames to 384×384 pixels, keeping other settings consistent with their original implementations.}
For \tao{all optical flow-dependent methods}, unified flow maps \tao{are} generated by RAFT~\cite{teed2020raft}. 
We do not apply data augmentation, as it may disrupt temporal consistency. 
The evaluation metrics \tao{we chose} include Intersection over Union (IoU$\uparrow$), F-measure ($F_\beta\uparrow$), Mean Absolute Error (MAE$\downarrow$), Balanced Error Rate (BER$\downarrow$) and Accuracy (ACC$\uparrow$).

\renewcommand{\tabcolsep}{1pt}
\renewcommand\arraystretch{1}
\begin{table*}[htbp]
  \centering 
  \small
  \begin{tabular}{clccccccccccccc} 
  \toprule 
  \multirow{2.45}{*}{\centering Task} &\multirow{2.45}{*}{Method} &\multirow{2.45}{*}{Year} &\multicolumn{5}{c}{VGSD-D~\cite{liu2024VGSDNet}}&&\multicolumn{5}{c}{MVGD-D (Ours)} &\multirow{2.45}{*}{Time(ms)}\\
  \cmidrule{4-8}  \cmidrule{10-14}
  &&&IoU$\uparrow$ &$F_\beta$$\uparrow$ &MAE$\downarrow$ &BER$\downarrow$ &ACC$\uparrow$ &&IoU$\uparrow$ &$F_\beta$$\uparrow$ &MAE$\downarrow$ &BER$\downarrow$ &ACC$\uparrow$ \\
  \midrule 
  \multirow{1}{*}{SOD}
  &MINet\cite{pang2020MINet} &CVPR'20 &71.84 &81.94 &0.162 &0.157 &0.869&&71.29 &81.29 &0.152 &0.132 &0.885&7.33 \\
  \midrule 
  \multirow{3}{*}{VSOD}
  &FSNet\cite{ji2021FSNet} &ICCV'21 &69.88 &78.22 &0.188 &0.179 &\SecondBest{0.914}&&66.90 &74.88 &0.190 &0.190 &\Best{0.938} &11.62\\
  &ISomer\cite{yuan2023ISomer} &ICCV'23 &77.58 &85.18 &0.127 &0.129 &0.905&&77.25 &85.81 &0.126 &0.114 &0.892 &16.14\\
  &UFO\cite{su2023UFO} &TMM'23 &64.69 &74.49 &0.235 &0.228 &0.873&&67.05 &77.16 &0.211 &0.192 &0.885 &22.28\\
  \midrule
  \multirow{1}{*}{SS}
  &SAM2\cite{ravi2024sam2} &arXiv'24 &78.60 &\SecondBest{89.28} &0.131 &0.125 &0.844&&78.18 &88.25 &0.121 &0.113 &0.841 &84.88\\
  \midrule
  \multirow{3}{*}{GSD} 
  &GSDNet\cite{lin2021GSDNet} &CVPR'21 &78.19 &86.46 &0.116 &0.111 &0.902&&76.34 &85.50 &0.125 &0.111 &0.894 &58.40\\
  &RFENet\cite{Fan2023RFENet} &IJCAI'23 &79.21 &88.60 &0.109 &0.105 &0.910&&76.45&87.29 &0.124 &0.109 &0.883 &39.14\\
  &GhostingNet\cite{yan2025GhostingNet} &TPAMI'24 &80.40 &88.83 &0.100 &\SecondBest{0.093} &0.905&&\SecondBest{80.01} &\SecondBest{88.75} &\SecondBest{0.104} &\SecondBest{0.094} &0.915 &32.94\\

  \midrule
  \multirow{1}{*}{VGSD} 
  &VGSDNet\cite{liu2024VGSDNet} &AAAI'24 &\SecondBest{80.72} &88.57 &\SecondBest{0.099} &0.096 &0.898 &&77.27 &85.59 &0.126 &0.110 &0.904&72.44\\
  \midrule
  \multirow{2}{*}{VMD} 
  &VMD-Net\cite{lin2023VMDNet} &CVPR'23 &74.12 &84.35 &0.136 &0.133 &0.879&&70.74 &81.92 &0.155 &0.139 &0.865 &54.40\\
  &MG-VMD\cite{warren2024MGVMD} &CVPR'24 &76.56 &84.68 &0.125 &0.123 &0.912&&73.69 &83.74 &0.134 &0.122 &0.887 &190.04\\
  \midrule
  \multirow{1}{*}{VGSD} 
  &Ours &- &\Best{86.57} &\Best{92.53} &\Best{0.064} &\Best{0.061} &\Best{0.935} &&\Best{82.62} &\Best{89.14} &\Best{0.090} &\Best{0.087} &\SecondBest{0.930}&190.9\\
  \bottomrule 
  \end{tabular}
  \caption{Quantitative comparison on VGSD-D and MVGD-D. 
  \taore{\Best{Red} and \SecondBest{Cyan} denote the best and second-best values, respectively.}}
  \label{tab:results_table}
\end{table*}

\subsection{Comparisons}
We evaluate our method against $11$ state-of-the-art approaches, including $1$ SOD method (MINet~\cite{pang2020MINet}), $3$ VSOD methods (FSNet~\cite{ji2021FSNet}, ISomer~\cite{yuan2023ISomer} and UFO~\cite{su2023UFO}), $1$ Semantic segmentation(SS) method (SAM2~\cite{ravi2024sam2}), $3$ GSD methods (GSD~\cite{lin2021GSDNet}, RFENet~\cite{Fan2023RFENet} and GhostingNet~\cite{yan2025GhostingNet}), one VGSD method (VGSDNet~\cite{liu2024VGSDNet}), and $2$ VMD methods (VMDNet~\cite{lin2023VMDNet} and MGVMD~\cite{warren2024MGVMD}). 
As shown in Table~\ref{tab:results_table}, our method achieves superior performance in both datasets. Specifically, compared to the second-best \tao{performing} method VGSD-Net~\cite{liu2024VGSDNet} on VGSD-D, our method performs better by IoU: $7.20\%\uparrow$, $F_\beta$: $4.46\%\uparrow$, MAE: $35.35\%\downarrow$, BER: $36.45\%\downarrow$, and ACC: $4.12\%\uparrow$.
\lyw{Furthermore, despite sharing the same backbone as GhostingNet, our method surpasses it, \tao{showing notable improvements on MVGD-D, as} IoU: $3.26\%\uparrow$, $F_\beta$: $0.44\%\uparrow$, MAE: $13.46\%\downarrow$, BER: $7.45\%\downarrow$, and ACC: $1.63\%\uparrow$. Notably, FSNet obtains a higher ACC but a lower IoU in two datasets, as it \taore{relies heavily on optical flow and} tends to overdetect glass regions in most scenes. It can also be seen from the table that single-image-based GSD methods perform worse than video-based GSD methods in most cases, mainly because they cannot exploit temporal information \tao{for GSD}. Similarly, SOD methods perform worse than GSD methods, as glass surfaces typically lack obvious appearances and require effective cues.}


The qualitative visual results are shown in Fig.~\ref{fig:qualitative_comparison_with_colorbar_MVGSD}. \lyw{In the $1st$ \tao{scene}, optical flow \tao{map exhibits obvious motions of} reflections, but networks such as Isomer, RFENet, MGVMD, and VGSD fail in the \tao{top}-left region with weak reflections. In the $2nd$ and $6th$ scenes, the reflections on the glass are relatively weak in the indoor environment, the optical flow \tao{maps display} motion inconsistencies from the transmitted layer, allowing our network to achieve better detection. It is worth noting that SAM2 and Isomer are affected by the semantic information of the transmitted layer, resulting in \tao{limited} performance. In the $3rd$ scene, the presence of railings \tao{at} the bottom area creates a glass-like region. Compared with Isomer and MGVMD, our method \tao{refines} the optical flow map to avoid introducing erroneous information. In these scenes, the proposed method performs better than \tao{all} competing methods, as our method \tao{is able to} effectively exploit motion inconsistency \tao{for GSD task}.}

\begin{table}[ht]
\centering
\small
\begin{tabular}{lccccc}
\toprule
\textbf{Module} & IoU$\uparrow$ & $F_\beta\uparrow$ & MAE$\downarrow$ & BER$\downarrow$  & ACC$\uparrow$\\
\midrule
\textit{A.} BS\,+\,BD & 74.31 & 80.87 & 0.140 & 0.135& 0.905 \\
\textit{B.} BS\,+\,RAFT\,+\,BF\,+\,BD & 75.59 & 82.12 & 0.136 & 0.131& 0.908 \\
\textit{C.} BS\,+\,CMFM\,+\,BT\,+\,TSD & 79.80 & 86.33 & 0.109 & 0.107& 0.915\\
\textit{D.} BS\,+\,BF\,+\,TAM\,+\,TSD & 78.74 & 85.24 & 0.117 & 0.112& 0.915 \\
\textit{E.} BS\,+\,CMFM\,+\,TAM\,+\,BD & 80.08 & 86.58 & 0.104 & 0.101 & 0.922\\
\textit{F.} \textit{w/o} P & 80.36 & 86.93 & 0.107 & 0.098& 0.922\\
\textit{G.} BS\,+\,CMFM\,+\,TAM\,+\,TSD &\Best{82.62} & \Best{89.14} &\Best{0.090} &\Best{0.087}& \Best{0.930}\\
\bottomrule
\end{tabular}
\caption{Ablation study of our network. `BS' \tao{denotes} the Swin Transformer backbone. `BD' \tao{represents} a simple decoder used to ablate TSD.  
\tao{Temporal Attention Module (TAM) is the collective term for HGAM and TCAM}
`BT' \tao{represents} a basic temporal module replacing TAM, and `BF' \tao{represents} a basic fusion module replacing CMFM. `P' \tao{denotes} the initial glass mask prediction used to refine optical flow map.}
\label{tab:ablation_guidance}
\end{table}
\renewcommand{\newsubheight}{0.106}
\begin{figure}[ht]
    \renewcommand{\tabcolsep}{0.8pt}
    \renewcommand\arraystretch{0.6}
    \centering
        \begin{tabular}{ccccccccc}
                \includegraphics[height=\newsubheight\linewidth, keepaspectratio]{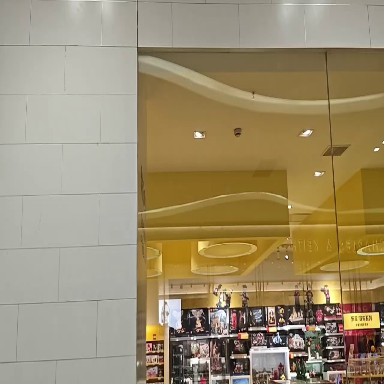}&
                \includegraphics[height=\newsubheight\linewidth, keepaspectratio]{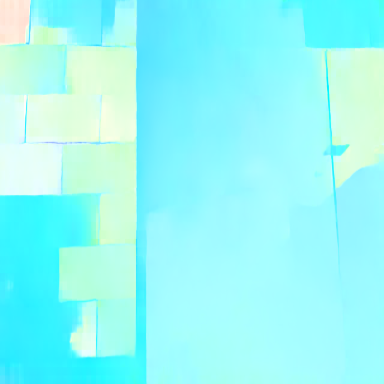}&
                \includegraphics[height=\newsubheight\linewidth, keepaspectratio]{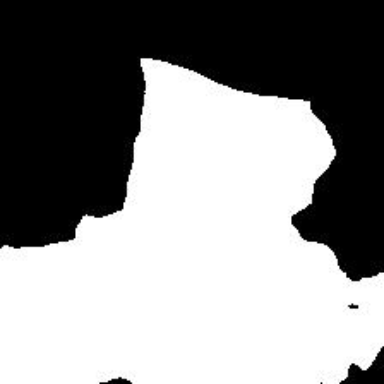}&
                \includegraphics[height=\newsubheight\linewidth, keepaspectratio]{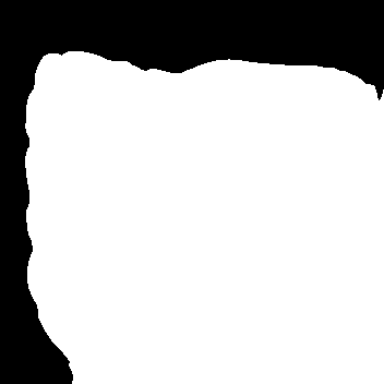}&
                \includegraphics[height=\newsubheight\linewidth, keepaspectratio]{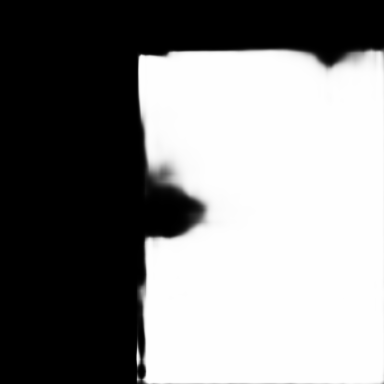}&
                \includegraphics[height=\newsubheight\linewidth, keepaspectratio]{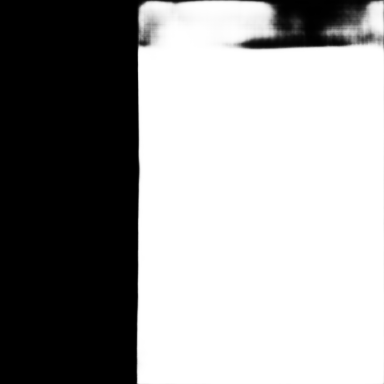}&
                \includegraphics[height=\newsubheight\linewidth, keepaspectratio]{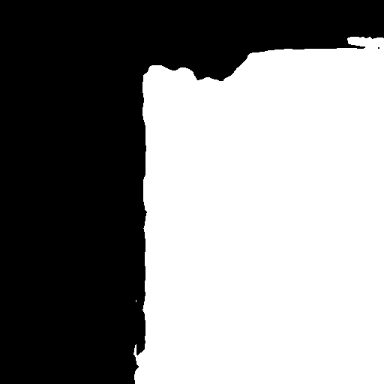}&
                \includegraphics[height=\newsubheight\linewidth, keepaspectratio]{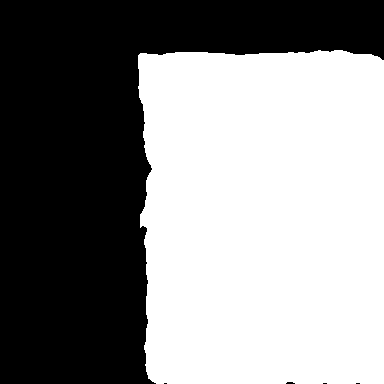}&
                \includegraphics[height=\newsubheight\linewidth, keepaspectratio]{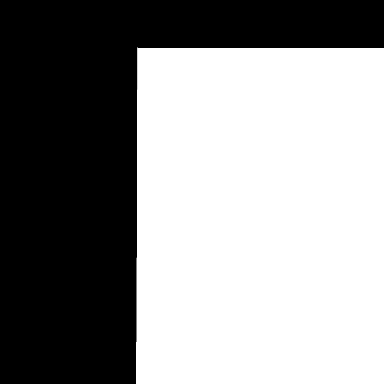}
                \\
  
                \includegraphics[height=\newsubheight\linewidth, keepaspectratio]{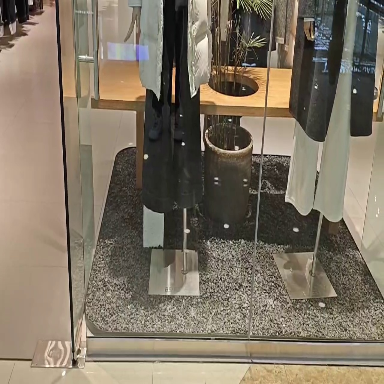}&
                \includegraphics[height=\newsubheight\linewidth, keepaspectratio]{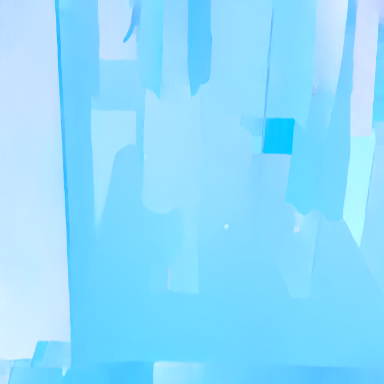}&
                \includegraphics[height=\newsubheight\linewidth, keepaspectratio]{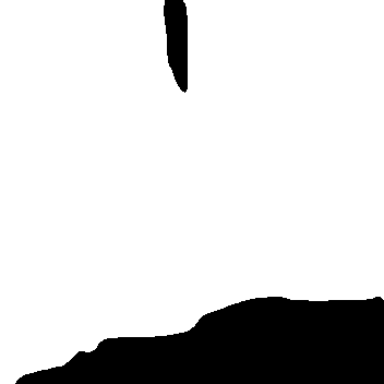}&
                \includegraphics[height=\newsubheight\linewidth, keepaspectratio]{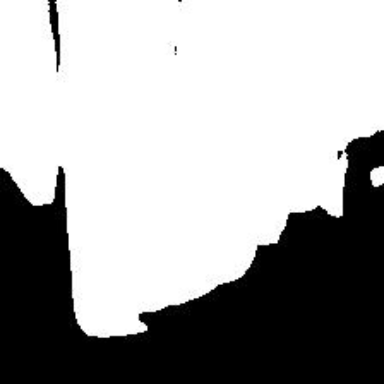}&
                \includegraphics[height=\newsubheight\linewidth, keepaspectratio]{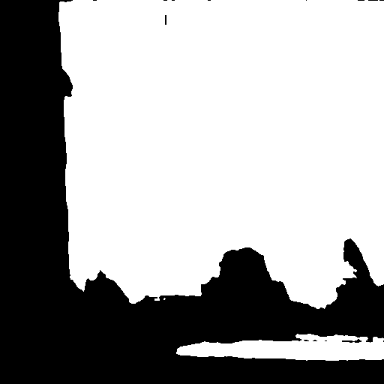}&
                \includegraphics[height=\newsubheight\linewidth, keepaspectratio]{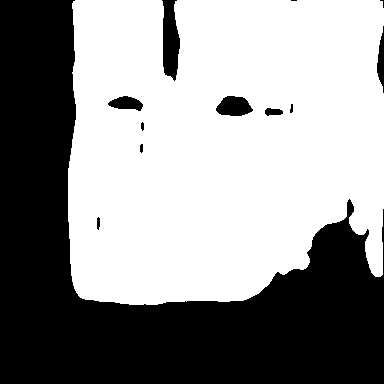}&
                \includegraphics[height=\newsubheight\linewidth, keepaspectratio]{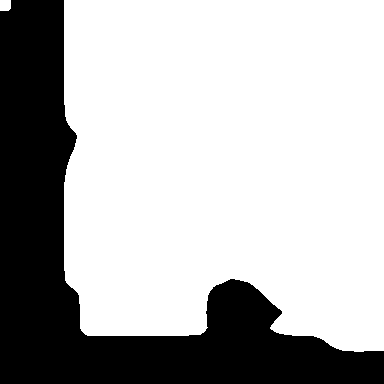}&
                \includegraphics[height=\newsubheight\linewidth, keepaspectratio]{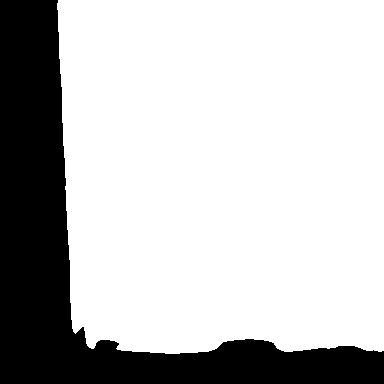}&
                \includegraphics[height=\newsubheight\linewidth, keepaspectratio]{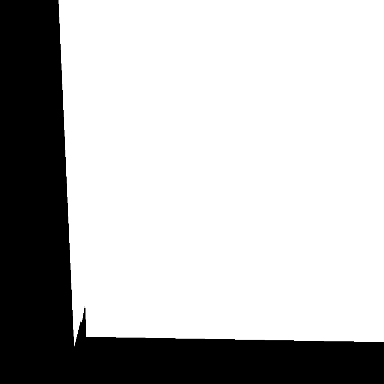}
                \\

                \small{Img}&
                \small{Flow}&
                \small{\textit{A}}&
                \small{\textit{B}}&
                \small{\textit{C}}&
                \small{\textit{D}}&
                \small{\textit{E}}&
                \small{Ours}&
                \small{GT}
                \\
                
        \end{tabular}
    \caption{The visual comparison of different ablated models.
     }
    \label{fig:ab_visual_result}
\end{figure}
\subsection{Ablation Study}

We report ablation results on MVGD-D in Tab.~\ref{tab:ablation_guidance}. 
The improvement of \textit{B} over \textit{A} on IoU demonstrates the effectiveness of motion cues in guiding GSD. The comparison between \textit{D} and \textit{G} indicates that CMFM brings a significant improvement to the network. Moreover, it can be seen from \textit{F} that the mask $P_{N-1}$ plays a crucial role in improving the performance of our network.
Visual \tao{comparisons} are shown in Fig.~\ref{fig:ab_visual_result}. The comparison between B and A further confirms that motion cues can guide \tao{GSD}, though further optimization is still needed. The comparisons of Models C–E with Model G show that each proposed module contributes positively to the overall performance.

\renewcommand{\newsubheight}{0.15}
\begin{figure}[t]
    \renewcommand{\tabcolsep}{0.8pt}
    \renewcommand\arraystretch{0.6}
    \centering
        \begin{tabular}{cccccc}
                \includegraphics[height=\newsubheight\linewidth, keepaspectratio]{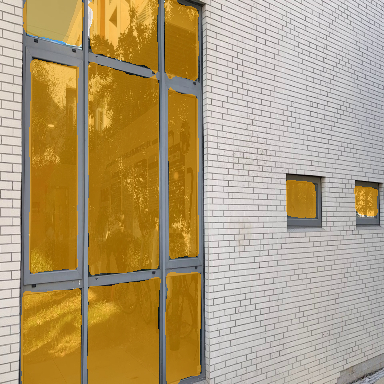}&
                \includegraphics[height=\newsubheight\linewidth, keepaspectratio]{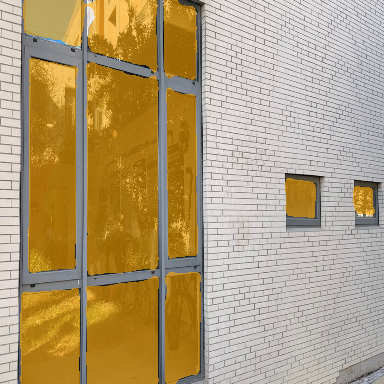}&
                \includegraphics[height=\newsubheight\linewidth, keepaspectratio]{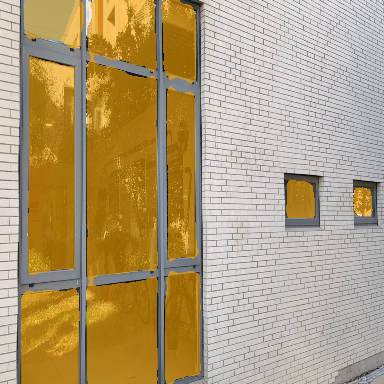}&
                \includegraphics[height=\newsubheight\linewidth, keepaspectratio]{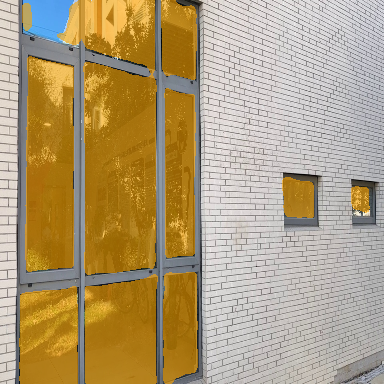}&
                \includegraphics[height=\newsubheight\linewidth, keepaspectratio]{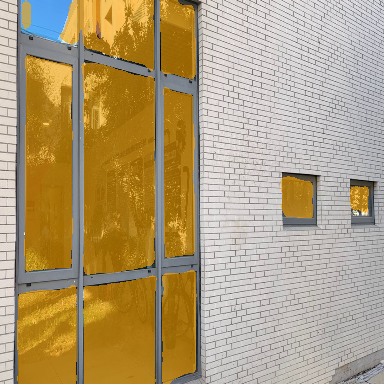}&
                \includegraphics[height=\newsubheight\linewidth, keepaspectratio]{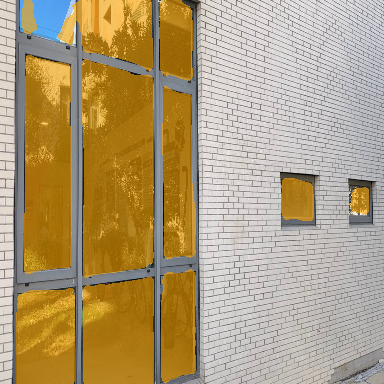}
                \\
  
                \includegraphics[height=\newsubheight\linewidth, keepaspectratio]{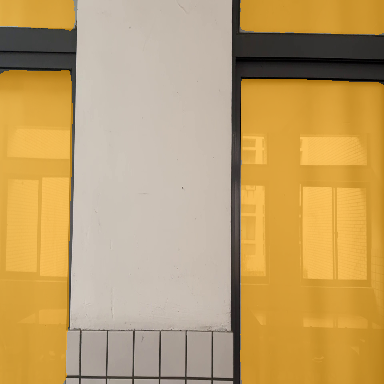}&
                \includegraphics[height=\newsubheight\linewidth, keepaspectratio]{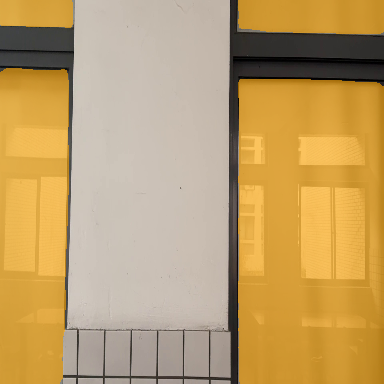}&
                \includegraphics[height=\newsubheight\linewidth, keepaspectratio]{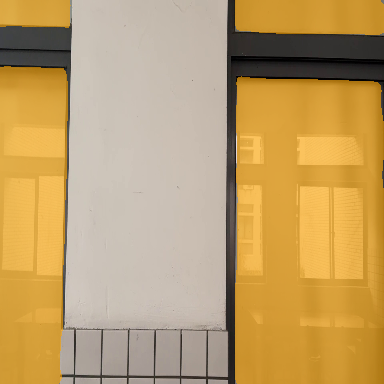}&
                \includegraphics[height=\newsubheight\linewidth, keepaspectratio]{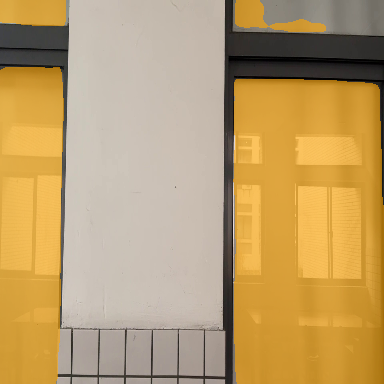}&
                \includegraphics[height=\newsubheight\linewidth, keepaspectratio]{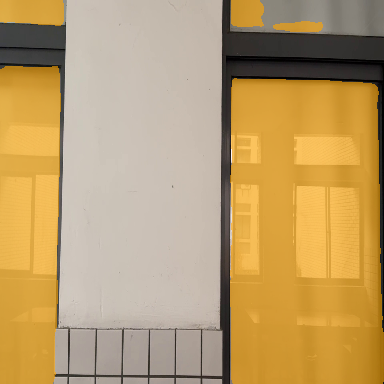}&
                \includegraphics[height=\newsubheight\linewidth, keepaspectratio]{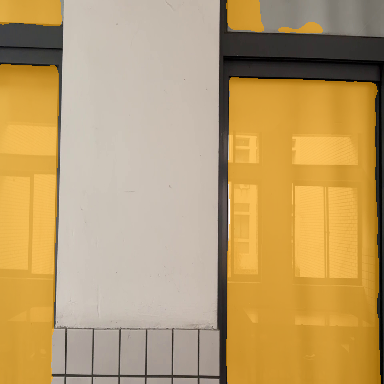}
                \\

                \fontsize{8.0pt}{\baselineskip}\selectfont{frame $t$}&
                \fontsize{8.0pt}{\baselineskip}\selectfont{frame $t+1$}&
                \fontsize{8.0pt}{\baselineskip}\selectfont{frame $t+2$}&
                \fontsize{8.0pt}{\baselineskip}\selectfont{frame $t+3$}&
                \fontsize{8.0pt}{\baselineskip}\selectfont{frame $t+4$}&
                \fontsize{8.0pt}{\baselineskip}\selectfont{frame $t+5$}
                \\
                
        \end{tabular}
    \caption{\tao{Typical} failure cases.}
    \label{fig:failure_cases}
\end{figure}

\section{Conclusion}
In this paper, we \tao{have proposed} a novel network, \tao{named} MVGD-Net, for \tao{VGSD} based on motion inconsistency cues. We \tao{have also proposed} three effective modules to exploit spatial and temporal features from videos to improve detection performance. Additionally, we \tao{have constructed} a large-scale dataset for \lywcr{VGSD}. Extensive experiments demonstrate that our method outperforms \tao{existing related} 
\lywcr{SOTA} methods.

Our method does have limitations, as it takes only three consecutive frames as input, which limits its ability to capture long-term temporal dependencies. As shown in Fig.~\ref{fig:failure_cases}, in the \tao{$1st$ scene}, the top-left glass region is missed in the $4th$ frame despite being consistently detected in previous frames. In the \tao{$2nd$ scene}, the top-right \tao{glass} region \tao{is under-}detected in the $4th$ frame. These failures highlight \tao{our} model's \tao{limitation in maintaining} consistent detection under appearance changes. \tao{Moreover, similar to existing image-based GSD methods, our method may mis-detect some open door/window, specifically \tao{glass-like} regions enclosed by their frames, as glass regions.}

\section*{Acknowledgments}
This work was supported by the National Natural Science Foundation of China (Grant No. 61902151).

\bibliography{ref}

\end{document}